%% file: main_arxiv.tex
\documentclass[10pt,twocolumn,letterpaper]{article}

\usepackage[accsupp]{axessibility} 
\usepackage[subtle,mathdisplays=normal,wordspacing=normal]{savetrees}
\usepackage{xcolor}

\newcommand{\camready}[1]{\textcolor{black}{~#1}}

\usepackage[pagenumbers]{cvpr} 

\input{preamble}
\usepackage{placeins}
\usepackage{adjustbox} 
\usepackage{longtable}
\usepackage{tocloft}

\makeatletter
\let\latexaddcontentsline\addcontentsline
\newcommand{\suppressmaintoc}{%
  \renewcommand{\addcontentsline}[3]{%
    \def\temp@ext{##1}%
    \def\toc@ext{toc}%
    \ifx\temp@ext\toc@ext
    \else
      \latexaddcontentsline{##1}{##2}{##3}%
    \fi
  }%
}
\newcommand{\restoretoc}{%
  \let\addcontentsline\latexaddcontentsline
}
\makeatother

\definecolor{cvprblue}{rgb}{0.21,0.49,0.74}
\usepackage[pagebackref,breaklinks,colorlinks,allcolors=cvprblue]{hyperref}

\def\paperID{6381} 
\def\confName{CVPR}
\def\confYear{2026}

\title{HandVQA: Diagnosing and Improving Fine-Grained Spatial Reasoning about Hands\\ in Vision-Language Models}

\author{
MD Khalequzzaman Chowdhury Sayem$^{1}$\thanks{Equal contribution.} \quad
Mubarrat Tajoar Chowdhury$^{1}$\footnotemark[1] \quad
Yihalem Yimolal Tiruneh$^{1}$ \\
Muneeb A. Khan$^{1}$ \quad
Muhammad Salman Ali$^{1}$ \quad
Binod Bhattarai$^{2,3,4}$\thanks{These authors jointly supervised this work.} \quad
Seungryul Baek$^{1}$\footnotemark[2] \\[3pt]
$^{1}$UNIST \quad
$^{2}$University of Aberdeen \quad
$^{3}$University College London \quad
$^{4}$Fogsphere (Redev.AI Ltd), UK
}

\begin{document}
\suppressmaintoc
\maketitle
\input{sec/abstract}
\input{sec/figures/main_teaser}
\input{sec/intro}
\input{sec/figures/pipeline_tex}
\input{sec/related_works}
\input{sec/dataset_cvpr}
\input{sec/tables/angle_distance_cvpr}
\input{sec/tables/rel_pos_cvpr}
\input{sec/experiments}

\input{sec/conclusion}

\input{acknowledge}

{
    \small
    \bibliographystyle{ieeenat_fullname}
    \bibliography{main}
}

\clearpage
\restoretoc
\onecolumn

\begin{center}
    {\Large \bfseries Supplementary Materials\par}
    \vspace{0.4em}
    {\Large \bfseries HandVQA: Diagnosing and Improving Fine-Grained Spatial Reasoning\\
    about Hands in Vision-Language Models\par}
    \vspace{1.2em}
    {\large \bfseries Contents\par}
\end{center}

\vspace{0.8em}

\renewcommand{\contentsname}{} 
\tableofcontents

\clearpage
\twocolumn

\input{supp_sec/dataset}
\input{supp_sec/experiments}


\end{document}

%% file: preamble.tex
\usepackage[utf8]{inputenc} 
\usepackage[T1]{fontenc}    
\usepackage{url}            
\usepackage{booktabs}       
\usepackage{amsfonts}       
\usepackage{nicefrac}       
\usepackage{microtype}      
\usepackage{xcolor}         
\usepackage[ruled,vlined]{algorithm2e}

\usepackage[table]{xcolor}
\definecolor{Gold}{RGB}{255,215,0}
\definecolor{Silver}{RGB}{192,192,192}
\definecolor{Bronze}{RGB}{205,127, 50}

\newcommand{\mub}[1]{\textcolor{black}{#1}}

\newcommand{\cvpr}[1]{\textcolor{black}{#1}}

\newcommand{\best}[1]{\cellcolor{Gold!70}\textbf{#1}}
\newcommand{\second}[1]{\cellcolor{Silver!150}#1}
\newcommand{\third}[1]{\cellcolor{Bronze!70}#1}

\newcommand{\first}[1]{\colorbox{Gold!70}{\textbf{#1}}}
\newcommand{\secondcaption}[1]{\colorbox{Silver!150}{\textbf{#1}}}
\newcommand{\thirdcaption}[1]{\colorbox{Bronze!70}{\textbf{#1}}}

\definecolor{darkgreen}{RGB}{0,100,0}

\newcommand{\greentext}[1]{\textcolor{darkgreen}{#1}}

\usepackage{amsmath}
\usepackage{array}
\usepackage{multirow}
\usepackage{makecell}
\usepackage{float}
\usepackage{caption}
\usepackage{graphicx}
\usepackage{cuted}
\usepackage{capt-of}
\usepackage{gensymb}
\usepackage{siunitx}
\usepackage{textcomp}
\usepackage{gensymb}
\usepackage{float}
\usepackage{subcaption}
\usepackage{soul}
\usepackage{tabularx}
\usepackage[subtle,mathdisplays=normal,wordspacing=normal]{savetrees}
\usepackage{enumitem}
\setlist{nosep}

\setlist{topsep=0pt, itemsep=0pt, partopsep=0pt, parsep=0pt}

\newcommand{\grayrule}{\arrayrulecolor{black!25}\midrule\arrayrulecolor{black}}

\usepackage{amsmath,amssymb}
\usepackage{tikz}
\usetikzlibrary{arrows.meta,positioning}

\definecolor{unistBlue}{HTML}{004C97}
\definecolor{unistTeal}{HTML}{0093D0}
\definecolor{unistGreen}{HTML}{00A78E}
\definecolor{unistGold}{HTML}{FFC20E}
\definecolor{moduleFill}{HTML}{F7F9FC}

\tikzset{
  module/.style = {draw=#1, very thick, rounded corners=3pt, fill=moduleFill,
                   minimum width=0.58\linewidth, minimum height=14mm, align=center},
  arrow/.style  = {-{Latex[scale=1]}, very thick},
  inside/.style = {font=\scriptsize}
}


%% file: sec/abstract.tex
\begin{abstract}
  Understanding the fine-grained articulation of human hands is critical in high-stakes settings such as robot-assisted surgery, chip manufacturing, and AR/VR-based human–AI interaction. Despite achieving near-human performance on general vision-language benchmarks, current vision-language models (VLMs) struggle with fine-grained spatial reasoning, especially in interpreting complex and articulated hand poses. We introduce HandVQA, a large-scale diagnostic benchmark designed to evaluate VLMs' understanding of detailed hand anatomy through visual question answering. Built upon high-quality 3D hand datasets (FreiHAND, InterHand2.6M, FPHA), our benchmark includes over 1.6M controlled multiple-choice questions that probe spatial relationships between hand joints, such as angles, distances, and relative positions. We evaluate several state-of-the-art VLMs (LLaVA, DeepSeek and Qwen-VL) in both base and fine-tuned settings, using lightweight fine-tuning via LoRA. Our findings reveal systematic limitations in current models, including hallucinated finger parts, incorrect geometric interpretations, and poor generalization. HandVQA not only exposes these critical reasoning gaps but provides a validated path to improvement. We demonstrate that the 3D-grounded spatial knowledge learned from our benchmark transfers in a zero-shot setting, significantly improving accuracy of model on novel downstream tasks like hand gesture recognition (+10.33\%) and hand-object interaction (+2.63\%). Project page, code, and dataset: \url{https://kcsayem.github.io/handvqa/}.
 
\end{abstract}

%% file: sec/figures/main_teaser.tex
\begin{figure}[t!]
\centering
\includegraphics[width=\linewidth]{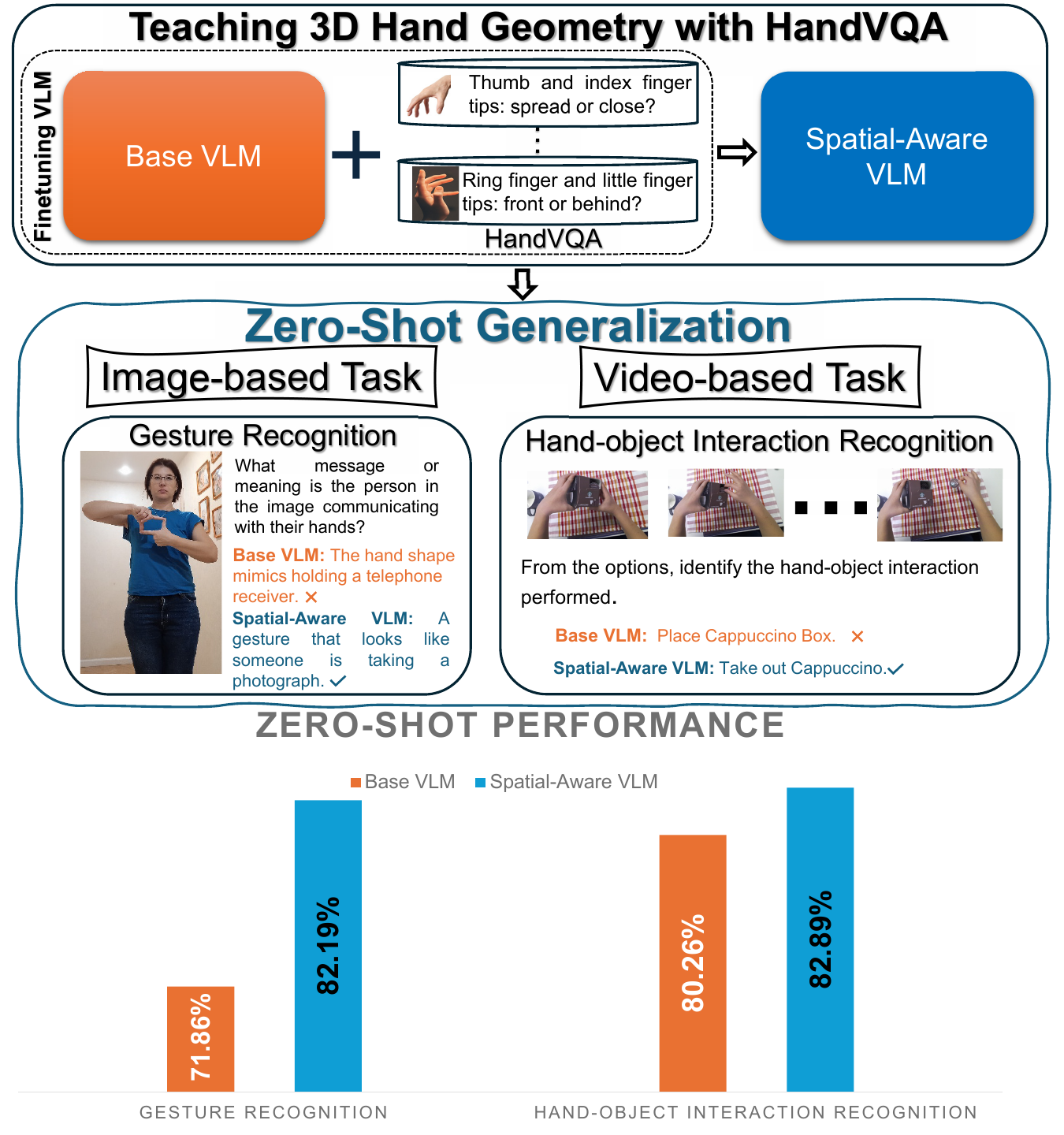}
\caption{ 
\textbf{Overview of HandVQA’s transfer effect.} Fine-tuning a base Vision-Language Model (VLM) on HandVQA teaches it explicit 3D hand geometry and joint-level spatial reasoning. The resulting Spatial-Aware VLM exhibits zero-shot generalization to novel downstream tasks: both image-based gesture recognition and video-based hand-object interaction recognition. Spatial-Aware VLM achieves consistent accuracy gains without task-specific training.
}
\label{fig:main_teaser}
\end{figure}

%% file: sec/intro.tex
\section{Introduction}
\input{sec/figures/intro_teaser}
Imagine a high-stakes environment like robot-assisted surgery, where a subtle misinterpretation of a surgeon’s hand gesture whether a finger is slightly bent or fully extended can mean the difference between a successful operation and a critical error \cite{ma2022surgical, ficuciello2021human}. Or consider chip manufacturing, where robotic agents mimic human motions with microscopic precision, and a misjudged fingertip angle can cause costly defects \cite{dui2023cost, doi:10.1243/0954405991517245}. In such scenarios, understanding the fine-grained articulation of the human hand is not just desirable, it is essential. \\
Vision Language Models (VLMs) have rapidly become the perceptual and decision-making core of embodied agents, AR/VR platforms, and multimodal assistants due to their ability to unify visual perception and natural-language reasoning within a single foundation model \cite{ma2024survey}. As these systems increasingly rely on such unified reasoning, fine-grained hand-pose understanding becomes crucial \citep{ xu2025egocentric, bao2024handsonvlm}. Hands are often the primary medium through which humans convey actions, intent, and control \citep{clough2020role}; misinterpreting whether a finger is curled, pointing, or relaxed can result in semantic confusion or physical failure. For VLMs to be safely deployed in these real-world contexts, robust, fine-grained hand understanding is a foundational requirement.\\
\noindent Despite significant advancements in general visual question answering (VQA), where state-of-the-art VLMs often achieve near-human accuracy on tasks like VQAv2 \citep{VQA}, they notably falter on tasks involving detailed spatial reasoning \citep{zhang2024vision, chen2025why, pothiraj2025capture, lee2025perspective}. Recent benchmarks highlight their particular weaknesses with basic spatial distinctions like left versus right, achieving only around 56\% accuracy compared to human performance at 99\% \citep{kamath2023s}. These shortcomings point to a reliance on superficial correlations rather than genuine geometric comprehension. Articulated hand poses, characterized by complex spatial relationships across 21 joints, remain a significant challenge.\\
\noindent To systematically address this challenge, we introduce \textit{HandVQA}, a diagnostic benchmark specifically designed to evaluate VLMs' fine-grained understanding of hand pose and anatomy through targeted visual question answering. HandVQA is constructed using precise 3D annotations from widely-used datasets—FreiHAND \citep{zimmermann2019freihand}, InterHand2.6M \citep{Moon_2020_ECCV_InterHand2.6M}, and FPHA \citep{garcia2018first}. We disentangle the hand pose estimation task into five separate subtasks: \mub{joint angles (measured at a single joint) and distances and relative positions along the X, Y, and Z axes (measured between pairs of joints)} (see Fig. \ref{fig:teaser}). While we draw conceptual motivation from \citet{delmas2022posescript}, our formulation is tailored specifically to the spatial nuances of hand anatomy. Our pipeline generates controlled multiple-choice questions probing these specific joint angles, distances, and relative positions, thereby minimizing ambiguity and encouraging genuine spatial reasoning (e.g., \textit{"Is the distal interphalangeal joint of the middle finger closer to that of the ring finger or the index finger?"}). By mapping 3D hand joint relations into structured natural language, HandVQA enables a diagnostic view into how well VLMs grasp spatial pose concepts.\\
\noindent We evaluate several open-source large VLMs, including LLaVA \citep{liu2023llava}, Qwen-VL \citep{Qwen-VL}, and DeepSeek \citep{lu2024deepseekvl}. These models are fine-tuned using parameter-efficient LoRA adapters \citep{hu2022lora} and evaluated on held-out samples from the same datasets used for training. Our findings indicate significant performance gaps: base models generally perform poorly, often similar to or below random guess, particularly on distance-related questions, reflecting a fundamental lack of spatial grounding. While fine-tuning markedly improves performance, confirming that VLMs can acquire spatial awareness with sufficient data; significant limitations persist, especially regarding the intricate task of accurately interpreting joint angles.\\
\noindent Our detailed analysis reveals key insights: VLMs often settle for simplified answers (e.g., repeatedly predicting ``close'' for distances or ``slightly bent'' for angles) indicating superficial alignment rather than true understanding. Furthermore, we observe that performance superiority in one spatial reasoning task rarely generalizes across others, highlighting that current VLM architectures might require stronger vision encoders or full-model fine-tuning to robustly capture intricate spatial relationships.\\
\noindent The implications of our study extend beyond benchmarking. Accurate hand pose interpretation is crucial in robotics for safe human-robot interactions, in AR/VR for immersive user experiences, and in medical assistance for sterile and precise gesture-based device control \citep{duan2025aha, bao2024handsonvlm, ohkawa:cvpr23, yangi2025artificial}. HandVQA addresses a critical gap in existing benchmarks, but our work demonstrates it is more than just a diagnostic tool; it is a robust training resource. We hypothesize that by mastering the 3D-grounded spatial reasoning in HandVQA, models acquire a transferable skill. We validate this by showing that models trained with HandVQA achieve significant zero-shot generalization to novel, unseen downstream tasks, including static Gesture Recognition and dynamic, video-based Hand-Object Interaction (See Fig. \ref{fig:main_teaser}). HandVQA thus provides a concrete path to improving VLM spatial reasoning capabilities broadly. To summarize our contributions include:

\begin{itemize}
    \item  We introduce a large-scale benchmark with over 1.6 million questions, created via an automated pipeline that generates anatomically-grounded VQA queries about 3D hand joint angles, distances, and relative positions.

    \item  We provide a Comprehensive evaluation of leading VLMs (e.g., LLaVA, DeepSeek, Qwen-VL), revealing their systemic failures in base spatial grounding (often performing at or below random guess) and highlighting persistent challenges, such as interpreting fine-grained joint angles, even after fine-tuning.

    \item  We demonstrate HandVQA is more than a diagnostic tool; it is a robust training resource. We show that the 3D-grounded spatial knowledge learned from our benchmark \textbf{transfers in a zero-shot setting}, significantly improving performance on novel, unseen downstream tasks. This includes a \textbf{+10.33\%} absolute accuracy gain on static \textbf{Gesture Recognition} and a \textbf{+2.63\%} gain on video-based \textbf{Hand-Object Interaction}.

\end{itemize}

%% file: sec/figures/intro_teaser.tex
\begin{figure*}[t!]
\centering
\includegraphics[width=\linewidth]{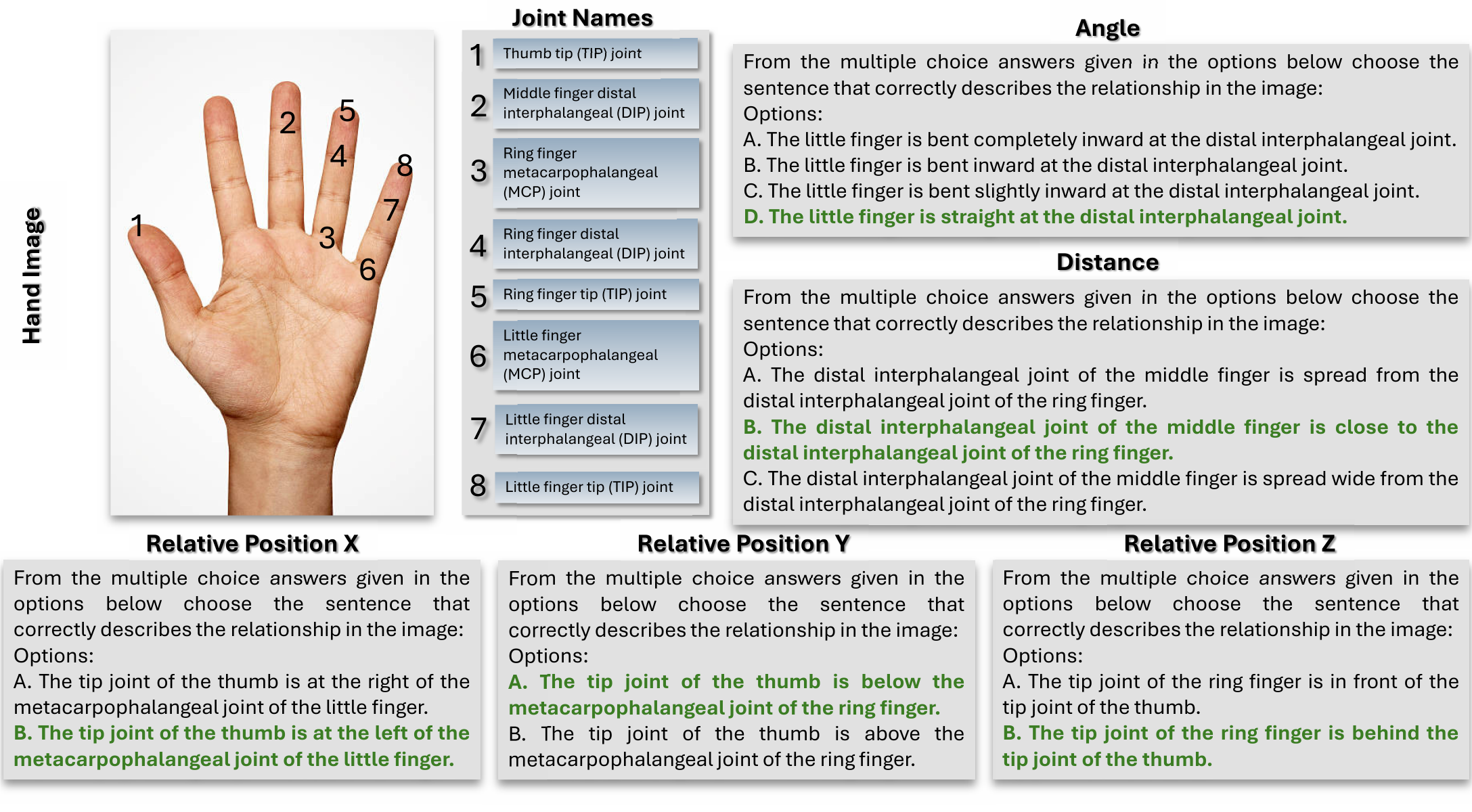}
\caption{\textbf{Overview of HandVQA Question Format.}
This figure illustrates the structure of our benchmark, which divides hand pose estimation into five sub-tasks: Angle, Distance, and Relative Position along X, Y, and Z axes. A hand image with annotated joint indices (top left) supports multiple-choice questions per task type, derived from 3D joint coordinates and the correct answers are shown in \greentext{\textbf{green}}.}
\label{fig:teaser}
\end{figure*}

%% file: sec/figures/pipeline_tex.tex
\begin{figure*}[t!]
\centering
\includegraphics[width=\linewidth]{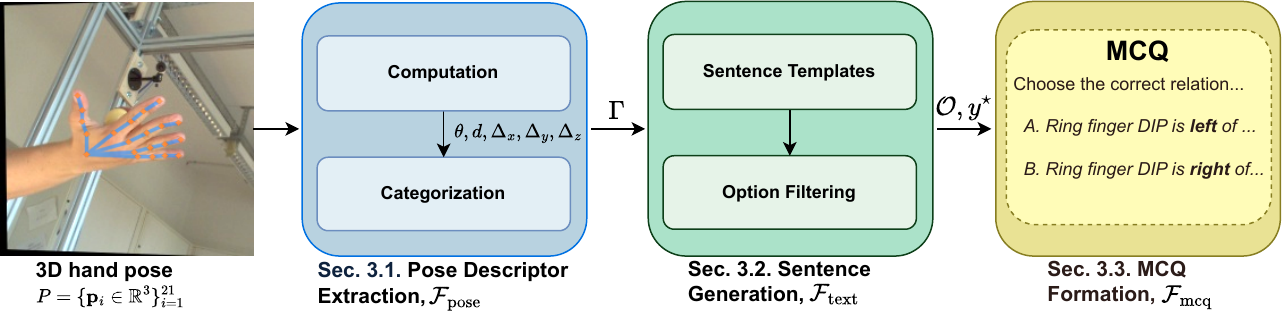}
\caption{\textbf{Overview of the HandVQA pipeline.}
The pipeline converts normalized 3D hand joints into interpretable VQA pairs through three deterministic stages:
(\textbf{1}) $\mathcal{F}_{\text{pose}}$ computes continuous pose descriptors: angles ($\theta$), distances ($d$), and relative positions ($\Delta_x, \Delta_y, \Delta_z$) and categorizes them into discrete pose descriptors ($\Gamma$);
(\textbf{2}) $\mathcal{F}_{\text{text}}$ fills deterministic sentence templates using $\Gamma$ and filters correct and incorrect options to form candidate answer options $(\mathcal{O})$ and the correct label \(y^\star\);
(\textbf{3}) $\mathcal{F}_{\text{mcq}}$ constructs multiple-choice questions (MCQ) by pairing each image with $\mathcal{O}$ and its correct label $y^\star$.}
\label{fig:handscript_pipeline}
\end{figure*}

%% file: sec/related_works.tex
\section{Related Works}
\textbf{Hallucinations and Alignment Failures in VLMs.}
Vision-language models (VLMs) often suffer from hallucinations, \ie, generating content not grounded in the image; such as nonexistent objects, incorrect attributes, or false spatial relations \cite{liu2024investigating}. For example, a model might describe a ring that isn’t there or misstate left/right positions \cite{li2023evaluating, kamath2023s}. \cvpr{Even state-of-the-art models} like GPT-4V \cite{achiam2023gpt} and BLIP-2 \cite{li2023blip} hallucinate objects \cite{li2023evaluating}, struggle with spatial terms \cite{kamath2023s}, or fabricate colors and shapes \cite{jiang2024hal}. These issues stem from weak visual-textual alignment and overreliance on language priors \cite{yang2025mitigating, jain2024neftune, wang2024mint}. \citet{liu2024survey} emphasizes the lack of grounding signals in training data. Recent solutions include evaluation frameworks such as POPE \cite{li2023evaluating}, NOPE \cite{lovenia-etal-2024-negative}, and Hal-Eval \cite{jiang2024hal}, as well as mitigation strategies, including contrastive decoding, reinforcement learning \cite{chen2024alleviating}, and robust visual encoders \cite{liu2025reducing}. Still, hallucinations remain prevalent, even in top-performing \cvpr{models \cite{mensink2023encyclopedic, NEURIPS2020_1457c0d6, chen2023pali, palm2023}} Our work exposes a distinct form of hallucination: \emph{pose hallucination}, where models misinterpret joint-level configurations. \cvpr{For instance, a model may predict a bent joint where the ground truth is straight, or repeatedly label finger pairs as ‘close’ despite substantial separation between them.} Such errors reveal fine-grained alignment failures and highlight the need for diagnostics beyond object-level VQA.

\noindent \textbf{Spatial and Relational Reasoning in Multimodal Models.}
Beyond identifying objects, true vision-language understanding requires reasoning about spatial and relational configurations. \citet{kamath2023s} revealed that many VLMs fail even simple spatial tasks like distinguishing a dog under a table versus on a table. Similarly, \citet{zhang2024sphere} presents SPHERE, a hierarchical evaluation of spatial skills from basic (positions, distances) to complex (occlusion, physical plausibility). On SPHERE, the best model achieved only 68\% accuracy far below human performance of 93\% \cite{zhang2024sphere}. These results show that high-level accuracy on generic VQA does not imply reliable spatial understanding \cite{shiri-etal-2024-empirical, pmlr-v235-wu24l}. \cvpr{To address these limitations, recent work has explored ways to explicitly inject spatial structure into multimodal models. For example,} \citet{yang2024improving} proposed a direction on improving model training by incorporating a spatial relation graph into vision-language pretraining, improving performance on reasoning tasks like VCR \cite{zellers2019recognition} and NLVR2 \cite{suhr2019corpus}. \citet{chen2024spatialvlm} proposed SpatialVLM, adding estimated depth maps and 3D spatial cues during training to overcome 2D limitations. Nonetheless, evaluating spatial reasoning remains difficult—many benchmarks conflate spatial skills with priors or world knowledge. \cvpr{Because VLMs \cite{Qwen-VL, lu2024deepseekvl, meta2024llama} frequently exploit shortcut cues in benchmarks such as CLEVR \cite{johnson2017clevr} and GQA \cite{hudson2019gqa}, their reported accuracy can obscure underlying spatial failures. This motivates the development of diagnostic evaluations like What’sUp \cite{kamath2023s} and SPHERE \cite{zhang2024sphere}.} Our HandVQA benchmark contributes by focusing on spatial relations within a single object: the human hand. Unlike benchmarks evaluating inter-object relationships, we target part–whole spatial structure, requiring understanding of joint kinematics and structured geometry. Since our questions are grounded in the real 3D coordinates, we test the model's ability to grasp Euclidean spatial concepts like \emph{distance} and \emph{angle}. 

\noindent \textbf{VQA Benchmarks and Fine-Grained Reasoning Evaluation.}
The VQA field has progressed from basic object/attribute Q\&A (e.g., VQAv1/v2) \cite{VQA} to compositional and fine-grained reasoning tasks. CLEVR \cite{johnson2017clevr} introduced logical queries in synthetic scenes; GQA \cite{hudson2019gqa} used real images and scene graphs for multi-step reasoning. Recent benchmarks like A-OKVQA \cite{schwenk2022okvqa} and Encyclopedic-VQA \cite{mensink2023encyclopedic} test domain knowledge, but fine-grained accuracy remains low: e.g., PaLI \cite{chen2023pali} achieves only 13\% on Encyclopedic-VQA \cite{mensink2023encyclopedic}. Our benchmark instead focuses on fine-grained spatial understanding. It aligns with works like NLVR2 \cite{suhr2019corpus} and CLEVR-3D \cite{yan2023comprehensive} that emphasize logical consistency and physical reasoning. \cvpr{Recent diagnostic frameworks such as VERIFY \cite{bi2025verify} evaluate if models follow reasoning chains rather than shallow cues.} II-MMR \cite{kil-etal-2024-ii} categorizes multi-hop VQA questions and shows many “hard” questions are solved via superficial features. To maintain focus, we keep HandVQA questions multiple-choice and relatively constrained—allowing clean assessment of spatial reasoning accuracy. It thus adds to diagnostic tools assessing whether progress in general VQA translates to true fine-grained understanding.

\noindent \textbf{Applications Requiring Precise Hand Pose Understanding.}
Precise hand articulation understanding is critical across domains. \cvpr{In robotics, accurate perception and manipulation of hand gestures is essential.} \citet{duan2025aha} introduce AHA, a VLM that detects and explains robotic manipulation failures, though it assumes accurate hand pose input. \citet{bao2024handsonvlm} present HandsOnVLM, forecasting future hand trajectories in egocentric video based on language queries. However, their method depends on reliable hand pose recognition \cite{bao2024handsonvlm}. Egocentric vision also relies on hand pose to infer intent. Ego4D emphasizes hand-object interactions \cite{xu2025egocentric}, while several works show that better 3D hand pose estimation improves activity recognition \cite{ismayilzada2025qort, ohkawa:cvpr23, cho2023transformer, reilly2025llavidal, on2025bigs, cha2024text2hoi}. More specifically, \citet{ohkawa:cvpr23} argue that hand pose is a compact representation of action. In augmented reality interfaces, devices like HoloLens or Meta Quest utilize hand tracking for gesture-based commands \cite{ohkawa:cvpr23}. Misinterpreting a finger’s position could be the difference between registering a “pinch” vs. a “point” gesture. Fine-grained evaluation like HandVQA can support the development of models that correctly interpret these nuances, reducing the risk of gesture misclassification (a form of hallucination in interaction). In surgical robotics or assistive tech, reliability is paramount: misclassification of a surgeon’s or disabled user’s gesture could have severe consequences \cite{yangi2025artificial}. Thus, many real-world applications demand detailed, error-free hand pose understanding. HandVQA offers a direct path to both benchmark and improve these capabilities. While many of the aforementioned works depend on reliable hand pose recognition as a prerequisite, our work demonstrates that the 3D-grounded spatial reasoning learned from HandVQA is a transferable skill. We validate this in our zero-shot experiments (Sec. \ref{sec:zero_shot_exp}), where models trained with HandVQA achieve significantly improved performance on these exact types of downstream tasks: static Gesture Recognition and dynamic, video-based Hand-Object Interaction. By forcing models to reason about joint-level spatial details, HandVQA not only evaluates but actively improves multimodal perception and grounding for these critical applications. 

%% file: sec/dataset_cvpr.tex
\section{Pipeline to generate HandVQA benchmark}
\label{sec:pipeline}

This section presents the design of the automatic VQA generation pipeline, which deterministically converts normalized 3D hand joint coordinates into natural language question-answer (QA) pairs for an image, $I$ given the normalized 3D hand pose \(P=\{\mathbf{p}_i\in\mathbb{R}^3\}_{i=1}^{21}\), and transforms geometric representations into linguistically interpretable forms. The three stages—pose descriptor extraction, sentence generation, and MCQ formation are described below.

\subsection{Pose Descriptor Extraction (\texorpdfstring{$\mathcal{F}_{\text{pose}}$}{F\_pose})}
Given normalized 3D hand joint coordinates 
\(P=\{\mathbf{p}_i\in\mathbb{R}^3\}_{i=1}^{21}\), $\mathcal{F}_{\text{pose}}$ computes continuous pose descriptors and categorizes them to describe the geometric and spatial relationships among hand joints. Formally, we define the continuous pose descriptor set as
\begin{equation}
    \Psi =
\begin{aligned}[t]
 &\Big\{\theta_j \mid j\in\mathcal{J}_\text{angle}\Big\} \cup\Big\{d_{(i,k)} \mid (i,k)\in\mathcal{J}_\text{pair}\Big\} \\[1mm]
 &\cup\Big\{\Delta_a(i,k) \mid (i,k)\in\mathcal{J}_\text{pair},\, a\in\{x,y,z\}\Big\}.
\end{aligned}
\end{equation}
where $\mathcal{J}_\text{angle}$ and $\mathcal{J}_\text{pair}$ denote the predefined joint index sets used for angle and pairwise computations, respectively. 
Here, indices $i,j,k$ refer to joint identifiers in the hand skeleton, and $a$ denotes one of the Cartesian axes $\{x,y,z\}$.

\begin{itemize}
    \item \textbf{Angle ($\theta_j$):} 
    Measures the bending degree at joint $j$, formed by its adjacent joints $a(j)$ and $b(j)$ along the same finger:
    \begin{equation}
    \theta_j = \arccos
    \frac{(\mathbf{p}_{a(j)}-\mathbf{p}_j)\cdot(\mathbf{p}_{b(j)}-\mathbf{p}_j)}
    {\|\mathbf{p}_{a(j)}-\mathbf{p}_j\|\,\|\mathbf{p}_{b(j)}-\mathbf{p}_j\|}.
    \end{equation}
    Angles are categorized into four bins based on fixed thresholds:
    \textit{bent completely inward}, \textit{bent inward}, \textit{bent slightly inward}, and \textit{straight}.
    \item \textbf{Distance ($d_{(i,k)}$):} 
    Quantifies the Euclidean distance between two joints \(i,k\):
    \(d_{(i,k)}=\|\mathbf{p}_i-\mathbf{p}_k\|_2.\)
    These distances are discretized into: \textit{close to}, \textit{spread from}, and \textit{spread wide from}.
    \item \textbf{Relative Position ($\Delta_a(i,k)$):} 
    Represents the signed offset of joint \(i\) relative to \(k\) along each axis:
    \(\Delta_a(i,k)=\langle \mathbf{p}_i-\mathbf{p}_k,\,\mathbf{a}\rangle.\)
    The relative positions are categorized as \textit{left/aligned/right}, \textit{below/aligned/above} and \textit{behind/aligned/in front of}. 
    
    Instances categorized as \textit{aligned} are excluded due to visual ambiguity. Further details in supplemental section 6.
\end{itemize}

\noindent
All threshold definitions are summarized in Table~\ref{tab:posecode_desc}. This process converts each continuous pose descriptor $\psi_n\in\Psi$ into a discrete pose descriptor label $\gamma_n$ using the category label from Table \ref{tab:posecode_desc}, forming the discrete pose descriptors
\begin{equation}
\Gamma = \{\gamma_n \mid \psi_n \in \Psi,\ n=1,2,\dots,N\},
\end{equation}
where $N$ denotes the total number of pose descriptors computed from a hand pose. This step ensures every continuous pose descriptor has a unique linguistic interpretation.
\input{sec/tables/category_threshold/category_threshold_single_column}

\subsection{Sentence Generation (\texorpdfstring{$\mathcal{F}_{\text{text}}$}{F\_text})}
Given discrete pose descriptors \(\Gamma\), the sentence generation step first fills deterministic templates with joint names and category labels to create natural language sentences \(\mathcal{S} = \{s_1, s_2, \dots ,s_n\}\).
Each descriptor type uses a fixed syntactic structure. For instance, the template for a distance descriptor is:
\begin{quote}
\textit{``The \underline{\{joint A\}} joint of the \underline{\{finger A\}} is \underline{\{category label\}} the \underline{\{joint B\}} joint of the \underline{\{finger B\}}.''}
\end{quote}
\noindent
If the category label is ``close to,'' the resulting sentence becomes:  
\textit{``The distal interphalangeal joint of the middle finger is close to the distal interphalangeal joint of the ring finger.''} \\
From the full set of generated sentences \(\mathcal{S}\), 
\(\mathcal{F}_{\text{text}}\) identifies the single sentence that corresponds  to the true category label and treats the remaining sentences for that joint pair as distractors in the option filtering step. This produces a pool of candidate answer options  \(\mathcal{O}\), along with the index of the correct label \(y^\star\),  which are later used to form an MCQ.

\subsection{MCQ Formation (\texorpdfstring{$\mathcal{F}_{\text{mcq}}$}{F\_mcq})}
Given the the set of candidate answer options \(\mathcal{O}\), and the index of the correct label \(y^\star\), \(\mathcal{F}_{\text{mcq}}\) constructs an MCQ prompt asking the model to identify the sentence that correctly describes the relationship between the specified joint(s).
To ensure diverse coverage of geometric relationships, multiple joint or joint–pair instances are considered for each descriptor type. Prior to filtering out ``aligned'' cases, up to 107 possible MCQs can be generated per image. For scalability, we randomly sample five joint or joint–pair instances per descriptor type, yielding a total of 25 MCQs per image across the five pose descriptor types (angle, distance, and relative positions along X/Y/Z). See Fig.~\ref{fig:teaser} for examples and Section~6 of the supplemental for sampling details.

\noindent
Through this process, the automatic VQA pipeline can generate descriptions for a vast number of poses in a fraction of the time it would take for manual annotations.

%% file: sec/tables/category_threshold/category_threshold_single_column.tex
\newcolumntype{L}[1]{>{\raggedright\arraybackslash}p{#1}}
\newcolumntype{C}[1]{>{\centering\arraybackslash}p{#1}}
\newcolumntype{R}[1]{>{\raggedleft\arraybackslash}p{#1}}

\newlength{\RowH}
\setlength{\RowH}{1.12\baselineskip}

\begin{table}[t!]
\caption{Pose descriptor categorization with conditions and illustration.}
  \centering
  \footnotesize
  \setlength{\tabcolsep}{0.8pt}
  \renewcommand{\arraystretch}{1.12}

  \begin{tabular}{@{} C{0.19\columnwidth} L{0.20\columnwidth} L{0.32\columnwidth} R{0.28\columnwidth} @{}}
    \toprule
    \textbf{Illustration} & \textbf{Pose Descriptor} & \textbf{Category Label} & \textbf{Condition} \\
    \midrule

    \multirow[c]{4}{*}{%
      \includegraphics[width=\linewidth,height=4\RowH,keepaspectratio]{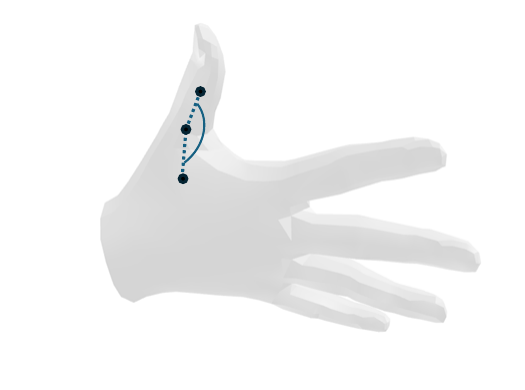}%
    }
      & \multirow[t]{4}{*}{\textbf{angle}}
      & bent completely inward & $\theta < 105^\circ$ \\
    & & bent inward            & $105^\circ \le \theta < 150^\circ$ \\
    & & bent slightly inward   & $150^\circ \le \theta < 170^\circ$ \\
    & & straight               & $\theta \ge 170^\circ$ \\
    \addlinespace[1pt]
    \grayrule

    \multirow[c]{3}{*}{%
      \includegraphics[width=\linewidth,height=3\RowH,keepaspectratio]{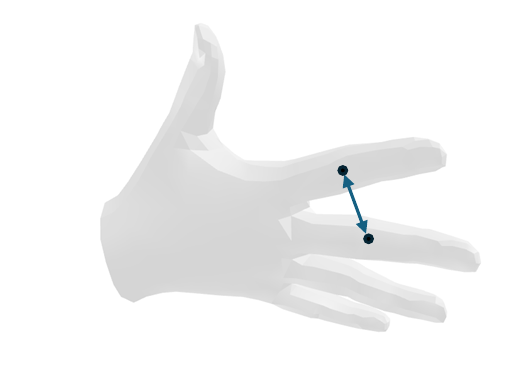}%
    }
      & \multirow[t]{3}{*}{\textbf{distance}}
      & close to         & $d < 0.1$ \\
    & & spread from      & $0.1 \le d < 0.3$ \\
    & & spread wide from & $d \ge 0.3$ \\
    \addlinespace[1pt]
    \grayrule

    \multirow[c]{3}{*}{%
      \includegraphics[width=\linewidth,height=3\RowH,keepaspectratio]{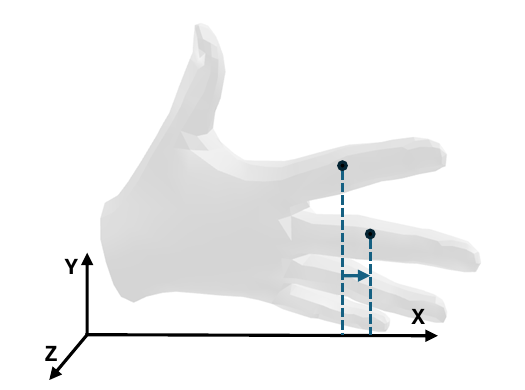}%
    }
      & \multirow[t]{3}{*}{\textbf{rel.\ pos.\ (X)}}
      & at the left of  & $\Delta_x < -0.15$ \\
    & & aligned         & $-0.15 \le \Delta_x < 0.15$ \\
    & & at the right of & $\Delta_x \ge 0.15$ \\
    \addlinespace[1pt]
    \grayrule

    \multirow[c]{3}{*}{%
      \includegraphics[width=\linewidth,height=3\RowH,keepaspectratio]{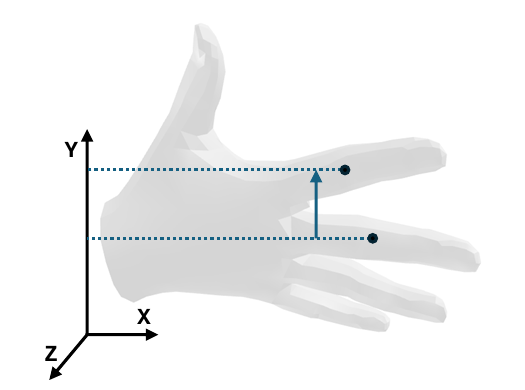}%
    }
      & \multirow[t]{3}{*}{\textbf{rel.\ pos.\ (Y)}}
      & below   & $\Delta_y < -0.15$ \\
    & & aligned & $-0.15 \le \Delta_y < 0.15$ \\
    & & above   & $\Delta_y \ge 0.15$ \\
    \addlinespace[1pt]
    \grayrule

    \multirow[c]{3}{*}{%
      \includegraphics[width=\linewidth,height=3\RowH,keepaspectratio]{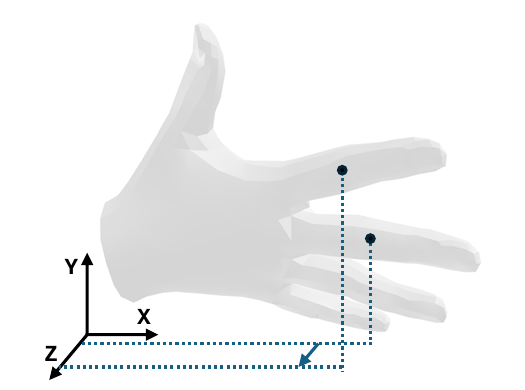}%
    }
      & \multirow[t]{3}{*}{\textbf{rel.\ pos.\ (Z)}}
      & behind      & $ \Delta_z < -0.15$ \\
    & & aligned     & $-0.15 \le \Delta_z < 0.15$ \\
    & & in front of & $\Delta_z \ge 0.15$ \\
    \bottomrule
  \end{tabular}

  \label{tab:posecode_desc}
\end{table}

%% file: sec/tables/angle_distance_cvpr.tex
\scriptsize
\setlength{\tabcolsep}{0.5pt}          
\renewcommand{\arraystretch}{0.95}      

\begin{table*}[!t]
\caption{Angle and Distance Results for all three models. The best, second-best, and third-best models for each dataset in each metric are highlighted as \first{Gold}, \secondcaption{Silver}, and \thirdcaption{Bronze}, respectively.}
  \centering
  \small
  
  \begin{tabular*}{\textwidth}{%
      @{\extracolsep{\fill}}
      lll
      @{\hspace{4pt}}
      rrrr
      @{\extracolsep{\fill}}
  }
    \toprule
    \multirow{2}{*}{Model} & \multirow{2}{*}{Tuned} & \multirow{2}{*}{Eval}
      & \multicolumn{2}{c}{Angle} & \multicolumn{2}{c}{Distance} \\
    \cmidrule(lr){4-5} \cmidrule(lr){6-7}
        &   &   & Accuracy $\uparrow$ & MAE $\downarrow$ & Accuracy $\uparrow$ & MAE $\downarrow$ \\
    \midrule \addlinespace

    \multicolumn{7}{@{}l}{\textbf{Base model (no tuning)}}\\
    DeepSeek Janus Pro 7B & – & InterHand2.6M & 34.10 & 0.883 & 45.55 & 0.657 \\
    DeepSeek Janus Pro 7B & – & FreiHAND  & 35.31 & 0.830 & 44.15 & 0.668 \\
    DeepSeek Janus Pro 7B & – & FPHA      & 26.46 & 0.991 & 39.02 & 0.819 \\

    \addlinespace
    \multicolumn{7}{@{}l}{\textbf{Finetuned Models}}\\
    DeepSeek Janus Pro 7B & InterHand2.6M & InterHand2.6M & \second{68.00} & \second{0.334} & \third{88.02} & \third{0.122} \\
    DeepSeek Janus Pro 7B & FreiHAND      & FreiHAND      & \second{61.30} & \second{0.402} & \second{85.23} & \second{0.151} \\
    DeepSeek Janus Pro 7B & FPHA          & FPHA          & \second{66.08} & \second{0.438} & \second{81.60} & \second{0.184} \\

    \midrule \addlinespace
    \multicolumn{7}{@{}l}{\textbf{Base model (no tuning)}}\\
    LLaVA Mistral 7B & – & InterHand2.6M & 40.08 & 0.739 & 16.20 & 1.293 \\
    LLaVA Mistral 7B & – & FreiHAND  & 42.48 & 0.678 & 13.18 & 1.342 \\
    LLaVA Mistral 7B & – & FPHA      & 23.38 & 1.011 & 13.57 & 1.353 \\

    \addlinespace
    \multicolumn{7}{@{}l}{\textbf{Finetuned Models}}\\
    LLaVA Mistral 7B & InterHand2.6M & InterHand2.6M & \best{74.35} & \best{0.263} & \best{90.79} & \best{0.094} \\
    LLaVA Mistral 7B & FreiHAND  & FreiHAND  & \best{62.91} & \best{0.382} & \best{86.19} & \best{0.141} \\
    LLaVA Mistral 7B & FPHA      & FPHA      & \best{68.37} & \best{0.401} & \best{83.99} & \best{0.161} \\

    \midrule \addlinespace
    \multicolumn{7}{@{}l}{\textbf{Base model (no tuning)}}\\
    Qwen 2.5 VL 7B Instruct & – & InterHand2.6M & 37.92 & 0.779 & 19.58 & 1.247 \\
    Qwen 2.5 VL 7B Instruct & – & FreiHAND  & 38.70 & 0.746 & 20.48 & 1.208 \\
    Qwen 2.5 VL 7B Instruct & – & FPHA      & 24.22 & 1.055 & 18.03 & 1.306 \\

    \addlinespace
    \multicolumn{7}{@{}l}{\textbf{Finetuned Models}}\\
    Qwen 2.5 VL 7B Instruct & InterHand2.6M & InterHand2.6M & \third{67.08}  & \third{0.341}  & \second{88.56}& \second{0.116}\\
    Qwen 2.5 VL 7B Instruct & FreiHAND    & FreiHAND      & \third{54.55}  & \third{0.483}  & \third{82.16}  & \third{0.182}  \\
    Qwen 2.5 VL 7B Instruct & FPHA        & FPHA          & \third{62.94}  & \third{0.481}  & \third{80.88}  & \third{0.192}  \\
    \midrule
  \end{tabular*}
  \label{tab:angle_distance_all_models}
\end{table*}
\normalsize

%% file: sec/tables/rel_pos_cvpr.tex
\scriptsize                    
\setlength{\tabcolsep}{5pt}  
\renewcommand{\arraystretch}{0.95}

\begin{table*}[!t]
\caption{Relative Position Results for all three models. The best, second-best, and third-best models for each dataset are highlighted as \first{Gold}, \secondcaption{Silver} and \thirdcaption{Bronze}, respectively.}
  \centering
  \small
  \begin{tabular*}{\textwidth}{@{}lll@{\hspace{4pt}}@{\extracolsep{\stretch{0.6}}}rrr@{}}
    \toprule
    \multirow{2}{*}{Model} & \multirow{2}{*}{Tuned} & \multirow{2}{*}{Eval}
      & \multicolumn{1}{c}{Rel.\ Pos.\ X}
      & \multicolumn{1}{c}{Rel.\ Pos.\ Y}
      & \multicolumn{1}{c}{Rel.\ Pos.\ Z} \\
    \cmidrule(lr){4-4} \cmidrule(lr){5-5} \cmidrule(lr){6-6}
        &   &   & Accuracy $\uparrow$ & Accuracy $\uparrow$ & Accuracy $\uparrow$ \\
    \midrule \addlinespace

    \multicolumn{6}{@{}l}{\textbf{Base model (no tuning)}}\\
    DeepSeek Janus Pro 7B & – & InterHand2.6M & 50.41 & 52.46 & 51.16 \\
    DeepSeek Janus Pro 7B & – & FreiHAND  & 49.80 & 51.55 & 50.03 \\
    DeepSeek Janus Pro 7B & – & FPHA      & 43.02 & 52.64 & 61.73 \\

    \addlinespace
    \multicolumn{6}{@{}l}{\textbf{Finetuned Models}}\\
    DeepSeek Janus Pro 7B & InterHand2.6M & InterHand2.6M & \third{92.58} & \third{96.40} & \third{92.16} \\
    DeepSeek Janus Pro 7B & FreiHAND      & FreiHAND      & \second{79.87}& \second{85.35}& \second{71.53}\\
    DeepSeek Janus Pro 7B & FPHA          & FPHA          & \third{89.94} & \third{86.45} & \second{88.12} \\

    \midrule \addlinespace
    \multicolumn{6}{@{}l}{\textbf{Base model (no tuning)}}\\
    LLaVA Mistral 7B & – & InterHand2.6M & 49.72 & 66.26 & 40.87 \\
    LLaVA Mistral 7B & – & FreiHAND  & 50.25 & 59.95 & 50.66 \\
    LLaVA Mistral 7B & – & FPHA      & 50.27 & 56.33 & 56.73 \\

    \addlinespace
    \multicolumn{6}{@{}l}{\textbf{Finetuned Models}}\\
    LLaVA Mistral 7B & InterHand2.6M & InterHand2.6M & \best{97.14} & \best{98.77} & \best{96.82} \\
    LLaVA Mistral 7B & FreiHAND  & FreiHAND  & \best{92.60} & \best{93.20} & \best{88.17} \\
    LLaVA Mistral 7B & FPHA      & FPHA      & \best{93.81} & \best{92.80} & \best{90.25} \\

    \midrule \addlinespace
    \multicolumn{6}{@{}l}{\textbf{Base model (no tuning)}}\\
    Qwen 2.5 VL 7B Instr. & – & InterHand2.6M & 48.98 & 49.78 & 49.33 \\
    Qwen 2.5 VL 7B Instr. & – & FreiHAND  & 49.17 & 49.60 & 50.19 \\
    Qwen 2.5 VL 7B Instr. & – & FPHA      & 50.98 & 48.53 & 49.79 \\

    \addlinespace
    \multicolumn{6}{@{}l}{\textbf{Finetuned Models}}\\
    Qwen 2.5 VL 7B Instr. & InterHand2.6M & InterHand2.6M & \second{94.90}& \second{97.49}& \second{94.11}\\
    Qwen 2.5 VL 7B Instr. & FreiHAND      & FreiHAND      & \third{76.67} & \third{80.12} & \third{70.23} \\
    Qwen 2.5 VL 7B Instr. & FPHA          & FPHA          & \second{93.45}& \second{90.61}& \third{87.63} \\
    \midrule
  \end{tabular*}
  \label{tab:relpos_all_models}
\end{table*}
\normalsize

%% file: sec/experiments.tex
\section{Experiments}
\label{experiments}

\subsection{Datasets, Models, and Evaluation Metrics}
\label{exp:model_eval_mat}
\cvpr{\textbf{Dataset Construction.}
We use three hand datasets to construct our HandVQA benchmark: FreiHAND \cite{zimmermann2019freihand}, InterHand2.6M \cite{Moon_2020_ECCV_InterHand2.6M}, and FPHA \cite{garcia2018first}. To ensure training and evaluations do not take too long, we create our own train/test splits by sampling a subset of images from the official splits of each dataset. The only exception is the FreiHAND test set, which we use in full due to its relatively small size. Further details on dataset construction are available in the supplemental section~7.}

\noindent \textbf{Models.} We evaluate four state-of-the-art 7B vision-language models—LLaVA Mistral\cite{liu2023llava}, DeepSeek Janus Pro \cite{lu2024deepseekvl}, and Qwen 2.5 VL Instruct \cite{Qwen-VL} on the HandVQA benchmark, using both base and LoRA \cite{hu2022lora} finetuned versions. Only 7B models were evaluated in this work due to GPU resource constraints. 

\noindent \textbf{Evaluation Metrics.} HandVQA comprises five sub-tasks derived from 3D hand joint annotations: angle, distance, and relative positions along the X, Y, and Z axes. For angle and distance, we report both accuracy and mean absolute error (MAE). While accuracy captures correct predictions, MAE reflects the average deviation from ground truth—crucial for ordinal categories where not all errors are equally severe (e.g., misclassifying      ``bent completely inward'' joint as ``straight'' is worse than as ``bent inward''). To compute MAE, we assign ordinal indices to each category based on increasing magnitude. For the angle task, the four categories—bent completely inward, bent inward, bent slightly inward, and straight—are mapped to class indices 0, 1, 2, and 3, respectively, reflecting increasing joint angles. For the distance task, the categories—close to, spread from, and spread wide from—are assigned indices 0, 1, and 2, corresponding to increasing joint distances. For relative position (X/Y/Z), we report accuracy only, as each is framed as a binary classification (e.g., left vs. right, below vs. above and behind vs. front). Ambiguous cases labeled “aligned” are excluded to ensure evaluation on clearly defined spatial relations.
\subsection{Results and Analysis}
\mub{Tables~\ref{tab:angle_distance_all_models} and \ref{tab:relpos_all_models} summarize the experimental results. In the following analysis, we examine the performance and behavior of the evaluated VLMs on our proposed HandVQA benchmark. Further detailed experimental analyses, including confidence analysis and cross-dataset comparisons, are provided in the supplementary (Sec.~8). }

\noindent \textbf{Scarcity of data the cause for VLMs' poor performance.} As per Table \ref{tab:angle_distance_all_models} and Table \ref{tab:relpos_all_models}, the base VLMs (base model without any finetuning) perform poorly on all pose descriptors, most often having an accuracy around random choice or worse. However, after fine-tuning, there are generally massive improvements across all metrics in all datasets for all the VLMs. This proves it is possible to train VLMs on spatial awareness of hands given abundant proper training data.\\
\noindent \textbf{VLMs struggle to grasp distance between joints.} As Table \ref{tab:angle_distance_all_models} shows, base VLMs generally seem to perform poorly on distance pose descriptor with LLaVA and Qwen performing well below the accuracy of 33.3\% accuracy that would have been achieved via random choice. Even the MAE remains high for two of these base models with the lowest MAE being 1.208 for Qwen on the FreiHAND dataset. While DeepSeek achieves an accuracy of more than random choice, it still remains low with the highest being 45.55\% on the InterHand2.6M dataset and the lowest MAE being a rather high 0.657 on the InterHand2.6M. The reasons for failure in case of base LLaVA and Qwen can be attributed to these models answering hand joints being ``close'' regardless of the situation. This is most severe in case of Qwen which answers ``close'' 93\% of the time when the actual answer is ``spread'' and 91.3\% of the time when the actual answer is ``spread wide'', as illustrated in the confusion matrix in supplemental. While base VLMs struggles to grasp the concept of distance between joints of fingers, for all the models across all datasets, the performance sees a massive boost upon fine-tuning, with the lowest accuracy being 80.88\% in case of Qwen fine-tuned on the FPHA dataset.\\
\noindent \textbf{VLMs struggle to grasp angle even after fine-tuning.} According to Table \ref{tab:angle_distance_all_models}, the performance of base VLMs across datasets excluding FPHA is generally substantially higher than the accuracy of 25\% that would have been achieved via random choice, with the lowest being 34.10\% for DeepSeek on the InterHand2.6M dataset. A common trend observed in the confusion matrix in the supplemental across all base models is that they all choose the option involving ``bent slightly inward'' in most cases irrespective of the actual answer. On the FPHA dataset, performance is significantly lower across all base VLMs compared to the other two datasets in terms of both accuracy and MAE, which can be attributed to FPHA being an egocentric dataset, indicating a bias in VLMs for allocentric viewing points. However, this trend is remedied after fine-tuning upon which FreiHAND is usually the dataset with the worst performance across all models. Unlike distance and relative position pose descriptors, where, upon fine-tuning, the accuracy generally jumps to above 80\%, in case of angles, however, the accuracy of fine-tuned model is below 70\% in most cases with the highest being 74.35\% for LLaVA fine-tuned on InterHand2.6M. Angle at joints of hands being a more intricate feature and being more representative of the pose of the hand means freezing the vision encoder, as is the case when fine-tuning with LoRA, becomes more of a limitation than for other pose descriptors. This can be overcome with a more powerful backbone or by fine-tuning the whole model on more data instead of fine-tuning with LoRA. Similar concerns have been raised in domains of VLMs expressing human-body pose \cite{feng2024chatpose}. \\
\noindent \textbf{Superiority in one task does not translate to superior performance in other tasks.} Among the base VLMs no model is superior to others across all tasks. In Table \ref{tab:angle_distance_all_models}, for angle, among base models, while base LLaVA on average performs best in terms of accuracy and base DeepSeek performs best in distance in terms of both MAE and accuracy. As can be seen in Table \ref{tab:relpos_all_models}, in case of Relative Position X, Y, and Z as well no base model dominates across all pose descriptors over other base models. However, upon finetuning, LLaVA comes out to be the superior model among all fine-tuned models for all pose descriptors across all metrics in all the datasets. \\
\noindent \textbf{Challenges in Interpreting Left/Right, Above/Below, and Front/Behind.}
Our results reveal that base VLMs lack a grounded understanding of fundamental spatial directions—left/right (X-axis), above/below (Y-axis), and front/behind (Z-axis). As shown in Table~\ref{tab:relpos_all_models}, the accuracy of all base models across datasets remains close to 50\%, effectively equivalent to random guessing in these binary classification tasks. This suggests that, without targeted adaptation, VLMs are unable to consistently reason about relative positions between joints. However, after fine-tuning on hand pose data, performance improves dramatically across all spatial axes. For example, LLaVA achieves about 97\% accuracy on all three axes in InterHand2.6M, and even lower-performing configurations exceed 70\% accuracy. These findings confirm that while VLMs do not possess inherent spatial grounding in directional concepts, they can acquire precise spatial reasoning abilities when exposed to sufficient task-specific supervision. This highlights the importance of our work, which explicitly isolates and evaluates these fine-grained spatial relations, enabling effective diagnosis and improvement of directional understanding in VLMs.

\subsection{Zero-Shot Generalization to Novel Hand-Related Tasks}
\label{sec:zero_shot_exp}
\input{sec/tables/zero_shot_exp}
Beyond diagnosing failures, we hypothesize that the 3D-grounded spatial reasoning taught by HandVQA is a transferable skill that should generalize to other novel, hand-related downstream tasks. To test this transferable learning, we designed a zero-shot generalization experiment, with results shown in Table \ref{tab:zero_shot_exp}. We selected two distinct, novel tasks for evaluation: static image-based Gesture recognition and video-based hand-object Interaction recognition. \cvpr{For these experiments, we chose LLaVA Mistral 7B \cite{liu2023llava}, best performing model in our benchmark and Qwen 2.5 VL 7B \cite{Qwen-VL}, given its support for temporal sequences (required for the video-based interaction task). Following common VLM finetuning practice \cite{feng2024chatpose, wang-etal-2024-demystifying, ShenHZ0LWCZFCVW24, liu2025llavaccontinualimprovedvisual}, we include a small, domain-agnostic instruction-alignment set alongside HandVQA (FreiHAND + InterHand2.6M + FPHA) to preserve general instruction-following capabilities during training.} 

\noindent \textbf{Gesture Recognition.} For this zero-shot task, we utilized the HaGRID dataset \cite{Kapitanov_2024_WACV}, which provides full-body images annotated with 34 hand gestures (e.g., `taking picture', `heart', `peace', etc.). \cvpr{To adapt this dataset into a VLM-evaluable format, we first convert each gesture label into a detailed natural-language description using Gemini \cite{gemini2024}. Then, based on these descriptions, we generate multiple-choice questions (MCQs) with one correct option and three carefully designed distractor options per sample.} Following this process, we constructed a final test set of 33,500 MCQs from the HaGRID test split (Further details is available in supplemental.). As shown in Table \ref{tab:zero_shot_exp}, the base LLaVA and Qwen models achieve 57.42\% and 71.86\% accuracy, respectively. After fine-tuning, their accuracies increase substantially to 69.58\% (LLaVA) and 82.19\% (Qwen), demonstrating strong positive transfer of 3D-grounded spatial reasoning to gesture understanding. Since both evaluated models show consistent improvement, we expect this trend to extend to other VLMs as well, given that HandVQA targets fundamental spatial reasoning skills shared across architectures.

\noindent \textbf{Hand-Object Interaction Recognition.} This experiment aimed to determine if the 3D spatial knowledge of hand geometry learned from HandVQA could transfer to improve the understanding of complex, egocentric hand-object interactions. We used the H2O dataset \cite{Kwon_2021_ICCV}, which provides diverse video-based interactions such as `taking out cappuccino from the coffee box',  `grabbing lotion', etc. From the H2O test split, we constructed a multiple-choice question (MCQ) benchmark, sampling 4 images per video for each question. As shown in Table \ref{tab:zero_shot_exp}, our results validate this hypothesis. The base Qwen-VL model achieved 80.26\% accuracy and the finetuned model achieved the highest accuracy of 82.89\%. LLaVA was not evaluated in this setting due to it's lack of temporal sequence support. This demonstrates that the spatial skills imparted by our benchmark are not limited to static poses but successfully transfer to enhance performance on dynamic, video-based interaction tasks.

%% file: sec/tables/zero_shot_exp.tex
\begin{table}[t]
\caption{
    \cvpr{\textbf{Zero-shot generalization results.} HandVQA fine-tuning strengthens 3D spatial reasoning, yielding higher accuracy (\%) on gesture and hand–object interaction recognition.}
  }
  \centering
  \begin{tabular*}{\columnwidth}{l @{\extracolsep{\fill}} cc}
  \toprule
  \textbf{Setup} & \makecell{\textbf{Interaction} \\ \textbf{Recognition}} & \makecell{\textbf{Gesture} \\ \textbf{Recognition}} \\
  \midrule
  LLaVA Mistral 7B \cite{liu2023llava} & -                & 57.42  \\
  Qwen 2.5 VL 7B \cite{Qwen-VL}      & 80.26                & 71.86 \\
  \midrule
  LLaVA Mistral 7B \textit{finetuned}      & -                &  69.58\\
  Qwen 2.5 VL 7B \textit{finetuned}      & \textbf{82.89}                & \textbf{82.19} \\
  \bottomrule
\end{tabular*}
  \label{tab:zero_shot_exp} 
\end{table}

%% file: sec/conclusion.tex
\section{Conclusion and Future Direction}

HandVQA provides the first large-scale benchmark for diagnosing fine-grained 3D spatial reasoning about hands in VLMs. Our analyses reveal that current models, despite strong general perception and instruction-following abilities, systematically fail on geometric relations such as joint angles, distances, and relative positions. By introducing a fully automated 3D-grounded annotation pipeline, we make these failures explicit and measurable. Furthermore, our zero-shot experiments show that HandVQA acts as an effective source of transferable 3D knowledge. Fine-tuning on our benchmark substantially improves performance on unseen tasks, such as gesture recognition and hand–object interaction.

\noindent \textbf{Future direction.} As the first large-scale benchmark for fine-grained hand-centric spatial reasoning, HandVQA focuses on core capabilities and naturally opens several research avenues: (i) Its discretized geometric representation offers a controlled and deterministic evaluation setup, with fixed thresholds providing a simplified view of the continuous 3D space. Future work can develop adaptive or learned mappings that better capture perceptual distinctions and support richer geometric reasoning. (ii) The templated language ensures unambiguous supervision, yet expanding to more diverse phrasing, comparative expressions, and explanations could yield stronger linguistic grounding of 3D geometry. (iii) Since the benchmark currently targets static images, extending it to video would enable reasoning about motion cues and contact dynamics central to many interaction tasks. (iv) Integrating HandVQA with Vision Language Action (VLA) models offers a path toward improved grasp planning, dexterous control, and embodied manipulation.


\noindent We hope HandVQA facilitates the development of multimodal models with stronger geometric understanding and more reliable physical reasoning.

%% file: acknowledge.tex
\section*{Acknowledgments}
This work is supported by NRF grant (No. RS-2025-00521013 50\%, No. RS-2025-02216916 10\%) and IITP grants (No. RS2020-II201336 Artificial intelligence graduate school program (UNIST) 5\%; No. RS-2025-25442824 AI Star Fellowship Program(UNIST) 5\%; No. RS-2022-II220264 Comprehensive video understanding and generation with knowledge-based deep logic neural network 10\%, No. RS-2025-25442149 LG AI STAR Talent Development Program for Leading Large-Scale Generative AI Models in the Physical AI Domain 10\%), funded by the Korean government (MSIT). This work is also supported by the InnoCORE program of the Ministry of Science and ICT(26-InnoCORE-01) 10\%.

%% file: supp_sec/dataset.tex
\section{Further HandVQA Pipeline Details}
\label{sec:further_dataset}
\input{sec/tables/list_of_pose_descriptors}
\input{supp_sec/dataset/mano_aligned_together}

In this study, we employ the following abbreviations: carpometacarpal (CMC), metacarpophalangeal (MCP), interphalangeal (IP), proximal interphalangeal (PIP), and distal interphalangeal (DIP). Figure \ref{fig:mano_joint_map}  illustrates the names and locations of the joints on the hand skeleton used in the generation pipeline for our HandVQA benchmark. Furthermore, Table \ref{tab:list_of_joints} provides a comprehensive list of 107 joints and joint pairs for which pose descriptors are calculated within the benchmark generation pipeline.


\subsection{Input to the HandVQA benchmark generation pipeline.}

In this section, we formally explain the input representation used by the HandVQA benchmark generation pipeline and define how raw 3D joint coordinates are normalized.

\noindent Let the raw 3D hand joint coordinates be denoted by
\begin{equation}
P^{\text{raw}}=\{\mathbf{p}^{\text{raw}}_i \in \mathbb{R}^3 \mid i=1,\dots,21\},
\end{equation}
where each joint is given by
\(\mathbf{p}^{\text{raw}}_i = (x_i, y_i, z_i)\).
The pipeline operates on a normalized version of these joints, producing
\[
P=\{\mathbf{p}_i \in \mathbb{R}^3 \mid i=1,\dots,21\},
\]
which is used throughout the HandVQA benchmark.

\subsubsection{Datasets with hand meshes}
For datasets providing full 3D hand meshes (\eg FreiHAND \cite{zimmermann2019freihand}, InterHand2.6M \cite{Moon_2020_ECCV_InterHand2.6M}), let
\begin{equation}
    V^{\text{raw}}=\{\mathbf{v}^{\text{raw}}_m \in \mathbb{R}^3 \mid m=1,\dots,M\}
\end{equation}
be the set of mesh vertices. We first center both vertices and joints using the mesh centroid
\[
\mathbf{c} = \frac{1}{M}\sum_{m=1}^M \mathbf{v}^{\text{raw}}_m,
\]
\[
\tilde{\mathbf{v}}_m = \mathbf{v}^{\text{raw}}_m - \mathbf{c}, \qquad
\tilde{\mathbf{p}}_i = \mathbf{p}^{\text{raw}}_i - \mathbf{c}.
\]
Let
\[
v^{\min}_a = \min_m \tilde{v}_{m,a}, \qquad
v^{\max}_a = \max_m \tilde{v}_{m,a},
\]
for each axis $a \in \{x,y,z\}$, and define the isotropic scale factor
\begin{equation}
    s = \frac{1}{\max_{a} \big(v^{\max}_a - v^{\min}_a\big)}.
\end{equation}


\subsubsection{Datasets without hand meshes}
Datasets such as FPHA~\cite{garcia2018first} does not provide hand meshes, so we compute the normalization using only the joint coordinates. The centroid and centered joints are
\[
\mathbf{c} = \frac{1}{21}\sum_{i=1}^{21} \mathbf{p}^{\text{raw}}_i,
\qquad
\tilde{\mathbf{p}}_i = \mathbf{p}^{\text{raw}}_i - \mathbf{c}.
\]
For each axis $a \in \{x,y,z\}$, let
\[
p^{\min}_a = \min_i \tilde{p}_{i,a}, \qquad
p^{\max}_a = \max_i \tilde{p}_{i,a},
\]
and define
\begin{equation}
    s = \frac{1}{\max_{a} \big(p^{\max}_a - p^{\min}_a\big)}.    
\end{equation}

\noindent The normalized joints are then
\begin{equation}
    \mathbf{p}_i = s\,\tilde{\mathbf{p}}_i.
\end{equation}

\noindent
This centering and isotropic scaling yields a normalized pose
\[
P = \{\mathbf{p}_i \in \mathbb{R}^3 \mid i=1,\dots,21\}
\]
that serves as the input to the pose descriptor extraction module $\mathcal{F}_{\text{pose}}$ in the HandVQA pipeline.


\subsection{Why cases with category label ``aligned'' in relative position are removed.}
In Figure \ref{fig:aligned_case}, while the little finger proximal interphalangeal joint (PIP) and the ring finger proximal interphalangeal joint (PIP) are occluded by the index finger and the middle finger, it can be deduced from the posture that the two PIP joints lie somewhere around the marked oval region, and it can also be deduced that the joints are close enough along the x-axis for the category label to be deemed as "aligned". While access to ground-truth joint coordinates allows us to ascertain their relative left-right relationship and generate corresponding MCQ data, the visual cue in itself is insufficient for a VLM to determine the relative left-right relationship of the two joints. The joints being too close along the x-axis makes their relationship ambiguous making it necessary to drop the relative position X information of the two joints when creating MCQ. Figure \ref{fig:aligned_case} shows an example of a scenario where aligned relative position makes the relationship ambiguous to interpret. Similarly in cases of all relative position pose descriptors, visual cues from cases where two joints are too close are deemed possibly ambiguous and dropped. 

\subsection{Sampling Details of Pose Descriptors and MCQs}

We describe here the sampling strategy used to generate a tractable yet comprehensive set of multiple-choice questions (MCQs) for each image. 
From the discrete pose descriptors produced by $\mathcal{F}_{\text{pose}}$, 
only descriptors corresponding to joints or joint pairs included in Table \ref{tab:list_of_joints} are considered for MCQ construction in HandVQA for scalibility. We define them as $\mathcal{T}$.

\paragraph{Angle descriptors.}
All anatomically valid finger-joint bending angles are used. Let
\[
\mathcal{J}_{\text{angle}}=\{j_1,\dots,j_{15}\}
\]
be the set of the 15 joints for which a bending angle is defined. For each $j \in \mathcal{J}_{\text{angle}}$, $\mathcal{F}_{\text{text}}$ generates four candidate sentences, one for each category label, and the correct sentence is selected.

\paragraph{Distance and relative-position descriptors.}
For distance and relative position along each axis ($x$, $y$, $z$), the total set of possible unordered joint pairs for each pose descriptor is
\[
\binom{21}{2}=210.
\]
However, many of these pairs are either anatomically implausible, rarely interacting with eachother in our daily activities, or redundant for fine-grained description. To ensure scalability, we restrict attention to the subset
\[
\mathcal{J}_{\text{pair}} \subseteq \{(i,k) \mid 1 \le i < k \le 21\}
\]
consisting only of joint pairs on \emph{adjacent fingers}. The exception is the thumb, which is permitted to compare with all other fingers because of its distinct opposable role in interacting with other fingers. This restriction yields a significantly smaller and semantically meaningful subset of pairs (shown in Table \ref{tab:list_of_joints}) for which $\mathcal{F}_{\text{text}}$ generates four candidate sentences, and the correct option.

\paragraph{Sampling.}
To maintain a manageable dataset size while ensuring descriptor diversity, we sample a fixed number of MCQs per image. Specifically, for each descriptor type \textit{Angle}, \textit{Distance}, and \textit{Relative Position X, Y, Z} we uniformly sample 5 distinct descriptor instances from the eligible joint or joint-pairs defined in $\mathcal{T}$ (demonstrated in Table \ref{tab:list_of_joints}). Each sampled element yields exactly one MCQ consisting of the prompt sentence and the option set $\mathcal{O}$ with a single correct answer. Therefore, each image results in 25 MCQs covering all five pose descriptor families.

\paragraph{Extensibility.}
Although the released benchmark samples from a reduced and anatomically meaningful subset $\mathcal{J}_{\text{pair}}$, it is possible to generate arbitrarily many MCQs per image within the limits of valid hand anatomy and application-specific needs by simple modifications.

\section{Dataset Statistics}
\label{sec:dataset_stat}
\input{supp_sec/dataset/word_cloud}
\input{supp_sec/dataset/train_eval_stat}
The details of each dataset and statistics of the benchmark are discussed in this section.  
\subsection{Dataset Construction}
We use three hand datasets to construct our \mub{HandVQA benchmark}: FreiHAND \cite{zimmermann2019freihand}, InterHand2.6M \cite{Moon_2020_ECCV_InterHand2.6M}, and FPHA \cite{garcia2018first}. Further details of them are discussed below,\\
\textbf{FreiHAND.} We construct our VQA training set using the last 30,000 images in the original training split of FreiHAND, which yields 742,575 VQA pairs consisting of all five pose descriptors. For the test set, we use the entire FreiHAND test split of size 3,960, yielding 98,261 VQA pairs consisting of all five pose descriptors.\\
\textbf{InterHand2.6M.} \mub{To construct training set, we use the 5 FPS version of the dataset and take images from the official training split of InterHand2.6M. We take images of subjects 5 to 26 in all right-hand postures, from the viewing point "cam400053" and "cam400064", yielding 132,999 VQA pairs from 5,348 images. The test split is also made up of images from the official training split of InterHand2.6M. We use images of subjects 1 to 4 in all right-hand postures with the images being from the same viewing points as our training split. This yields 97,806 VQA pairs from 3,934 images.}\\
\textbf{FPHA.} The training set is constructed using all video sequences of subjects 1,2,3, and 4 performing all the actions in the dataset, yielding 374,056 VQA pairs from 15,000 randomly selected images. The test set is constructed using video sequence 1 images of subjects 5 and 6 performing all the actions in the dataset, yielding 212,336 VQA pairs from 8,511 images.

\subsection{Balanced Coverage of Spatial Reasoning Tasks}
The Figure \ref{fig:train_eval_stat} presents the distribution of question types: Angle, Distance, and Relative Position (X, Y, Z axes) across the training and evaluation splits for each dataset used in HandVQA: FPHA \cite{garcia2018first}, FreiHAND \cite{zimmermann2019freihand}, and InterHand2.6M \cite{Moon_2020_ECCV_InterHand2.6M}.

In both the training (left) and evaluation (right) plots (as shown in Fig. \ref{fig:train_eval_stat}), each dataset exhibits a balanced distribution across all five question types. This uniformity ensures that no particular spatial reasoning category is over- or under-represented, facilitating fair comparison and comprehensive evaluation across models.

The proportions of each question type are consistent across all datasets, making HandVQA a well-structured benchmark for studying fine-grained multimodal understanding across diverse datasets.

\subsection{Word Cloud Analysis of Pose Descriptors}

Figure~\ref{fig:word_cloud} shows a word cloud visualization constructed from the textual pose descriptors used throughout the HandVQA benchmark. This visual highlights the most frequently occurring terms across the dataset's five question types—angle, distance, and relative positions in X, Y and Z axis.

Many of the most prominent words (e.g., \textit{interphalangeal joint}, \textit{tip joint}, \textit{metacarpophalangeal joint}, \textit{thumb}, \textit{index finger}) are directly tied to the anatomical joint names and relationships defined in our task design. As shown in Table~\ref{tab:list_of_joints}, these joint names form the core of the five types of pose descriptions used to generate structured language annotations.

Specifically:
\begin{itemize}
    \item \textbf{``Tip''} appears prominently because many distance and relative position comparisons involve tip joints, such as \textit{Thumb-Tip vs. Index-Tip} or \textit{Thumb-Tip vs. Ring-MCP}.
    
    \item \textbf{``Interphalangeal''} and its variations (e.g., proximal, distal) are common due to their presence in all pose descriptors as shown in Table \ref{tab:list_of_joints}.
    
    \item \textbf{Category label related terms} such as \textit{bent inward}, \textit{straight}, \textit{spread wide}, \textit{spread}, \textit{left}, \textit{behind}, and \textit{completely inward} come from the classification vocabulary used to describe joint relationships across all five pose descriptors.
    \item \textbf{Finger names} like \textit{thumb}, \textit{index finger}, \textit{ring finger}, and \textit{little finger} occur frequently because they are used systematically across all pose descriptor types.
\end{itemize}

This word cloud highlights the anatomical precision and task consistency of our benchmark's language component, demonstrating that the generated textual annotations are grounded in structured and meaningful joint relationships.

\subsection{Detailed Category Label Statistics}
\input{supp_sec/dataset/catagory_stats}
\camready{We further analyze the distribution of category labels within each pose descriptor family to better understand the underlying data characteristics. Table~\ref{tab:label_frequency_compact} summarizes the frequency of all discrete labels used across angle, distance, and relative position descriptors.}

\camready{For angle descriptors, the distribution is skewed toward mid-range articulation states, with \textit{bent slightly inward} (130,993) and \textit{bent inward} (107,143) dominating the dataset, while extreme configurations such as \textit{bent completely inward} (21,622) are comparatively underrepresented. The \textit{straight} category (74,007) occupies an intermediate proportion.}

\camready{For distance descriptors, the dataset is heavily biased toward larger separations between joints. The label \textit{spread wide from} (222,429) is the most frequent, followed by \textit{spread from} (109,353), whereas \textit{close to} (1,983).}

\camready{In the case of relative position descriptors, the distributions are relatively balanced but still exhibit mild asymmetries. Along the X-axis, \textit{at the left of} (198,988) appears more frequently than \textit{at the right of} (133,109). Similarly, for the Y-axis, \textit{above} (215,725) is more common than \textit{below} (112,984). Along the Z-axis, the distribution between \textit{in front of} (177,216) and \textit{behind} (152,481) is comparatively more balanced, though still slightly skewed.}

%% file: sec/tables/list_of_pose_descriptors.tex
\begin{table*}[t!]
\centering
\caption{List of joints/joint-pairs on which angles, distances, and relative positions in X,Y, and Z axes are calculated. A total of 107 different joints/joint-pairs across all pose descriptors are considered.}
\renewcommand{\arraystretch}{1.2} 
\setlength{\tabcolsep}{4pt} 
\begin{tabular}{p{0.28\textwidth} p{0.33\textwidth} p{0.33\textwidth}}
\hline
\textbf{Angle Pose Descriptors} & \textbf{Distance Pose Descriptors} & \textbf{Relative Position Pose Descriptors (XYZ)} \\ \hline
Thumb-MCP               & Thumb-MCP \textit{vs.} Index-PIP        & Thumb-MCP \textit{vs.} Index-PIP                       \\
Index-PIP               & Index-PIP \textit{vs.} Middle-PIP       & Index-PIP \textit{vs.} Middle-PIP                      \\
Middle-PIP              & Middle-PIP \textit{vs.} Ring-PIP        & Middle-PIP \textit{vs.} Ring-PIP                       \\
Ring-PIP                & Ring-PIP \textit{vs.} Little-PIP        & Ring-PIP \textit{vs.} Little-PIP                       \\
Little-PIP              & Thumb-Tip \textit{vs.} Index-Tip        & Thumb-Tip \textit{vs.} Index-Tip                        \\
Thumb-IP                & Index-Tip \textit{vs.} Middle-Tip       & Index-Tip \textit{vs.} Middle-Tip                      \\
Index-DIP               & Middle-Tip \textit{vs.} Ring-Tip        & Middle-Tip \textit{vs.} Ring-Tip                       \\
Middle-DIP              & Ring-Tip \textit{vs.} Little-Tip        & Ring-Tip \textit{vs.} Little-Tip                       \\
Ring-DIP                & Thumb-Tip \textit{vs.} Index-DIP       & Thumb-Tip \textit{vs.} Index-DIP                       \\
Little-DIP              & Thumb-Tip \textit{vs.} Middle-DIP       & Thumb-Tip \textit{vs.} Middle-DIP                      \\
Little-MCP              & Thumb-Tip \textit{vs.} Ring-DIP       & Thumb-Tip \textit{vs.} Ring-DIP                       \\
Ring-MCP                & Thumb-Tip \textit{vs.} Little-DIP         & Thumb-Tip \textit{vs.} Little-DIP                       \\
Middle-MCP              & Thumb-Tip \textit{vs.} Index-MCP       & Thumb-Tip \textit{vs.} Index-MCP                      \\
Index-MCP               & Thumb-Tip \textit{vs.} Middle-MCP        & Thumb-Tip \textit{vs.} Middle-MCP                       \\
Thumb-CMC               & Thumb-Tip \textit{vs.} Ring-MCP       & Thumb-Tip \textit{vs.} Ring-MCP                     \\
                        & Thumb-Tip \textit{vs.} Little-MCP         & Thumb-Tip \textit{vs.} Little-MCP                     \\
                        & Index-MCP \textit{vs.} Index-DIP       & Index-MCP \textit{vs.} Index-DIP                     \\
                        & Index-DIP \textit{vs.} Middle-DIP      & Index-DIP \textit{vs.} Middle-DIP                      \\
                        & Middle-DIP \textit{vs.} Ring-DIP         & Middle-DIP \textit{vs.} Ring-DIP                      \\ 
                        & Ring-DIP \textit{vs.} Little-DIP       & Ring-DIP \textit{vs.} Little-DIP                      \\    
                        & Thumb-Tip \textit{vs.} Middle-Tip        & Thumb-Tip \textit{vs.} Middle-Tip                                                  \\
                        & Middle-Tip \textit{vs.} Little-Tip       & Middle-Tip \textit{vs.} Little-Tip                                                  \\
                        & Index-Tip \textit{vs.} Ring-Tip        & Index-Tip \textit{vs.} Ring-Tip                                                    \\
\hline
\end{tabular}
\label{tab:list_of_joints}
\end{table*}

%% file: supp_sec/dataset/mano_aligned_together.tex



\begin{figure}[!t]
\centering
\includegraphics[width=\linewidth]{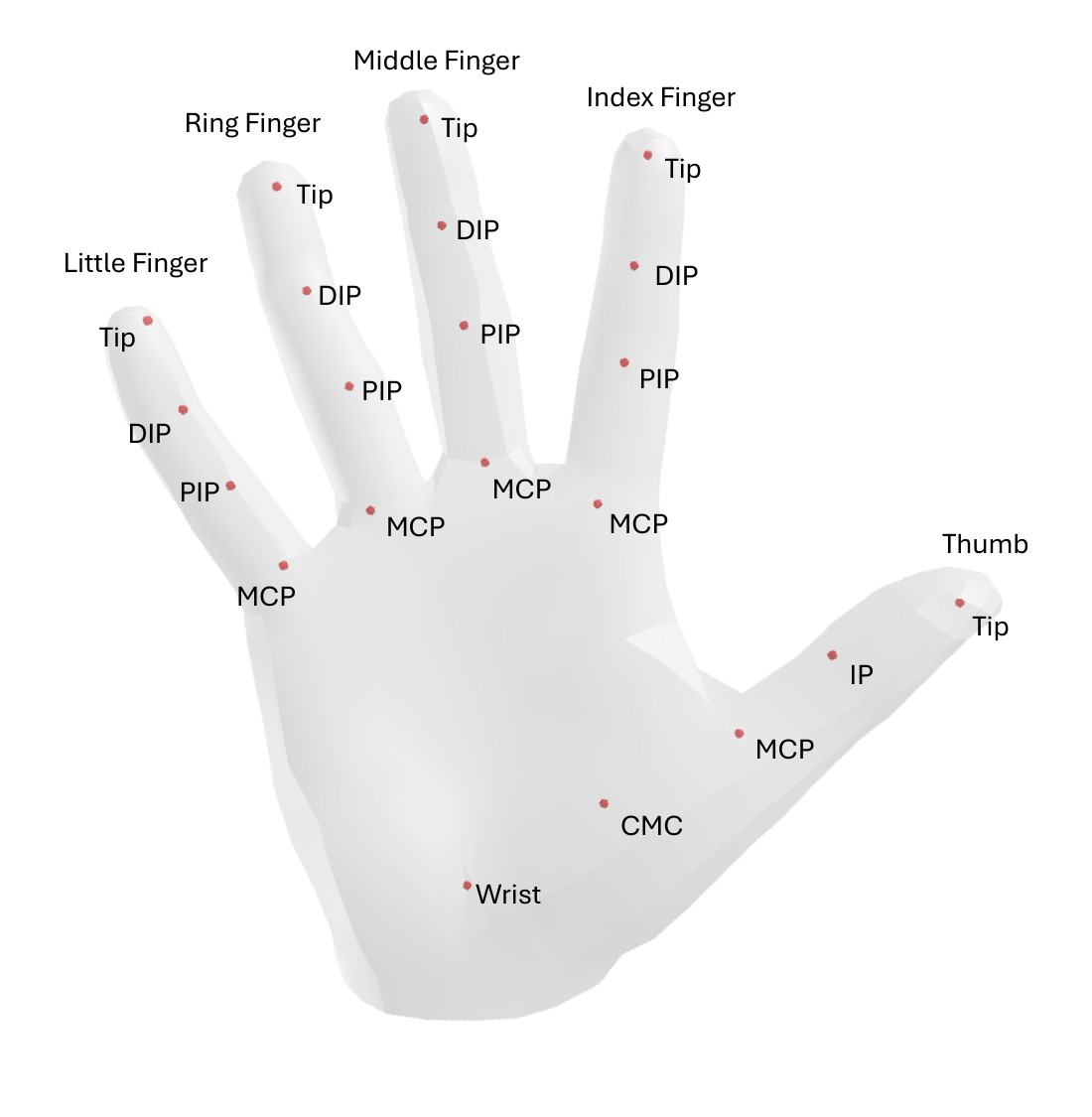}
\caption{The map of the hand skeleton used in our HandVQA benchmark generation pipeline.}
\label{fig:mano_joint_map}
\end{figure}

\begin{figure}[!t]
\centering
\includegraphics[width=\linewidth]{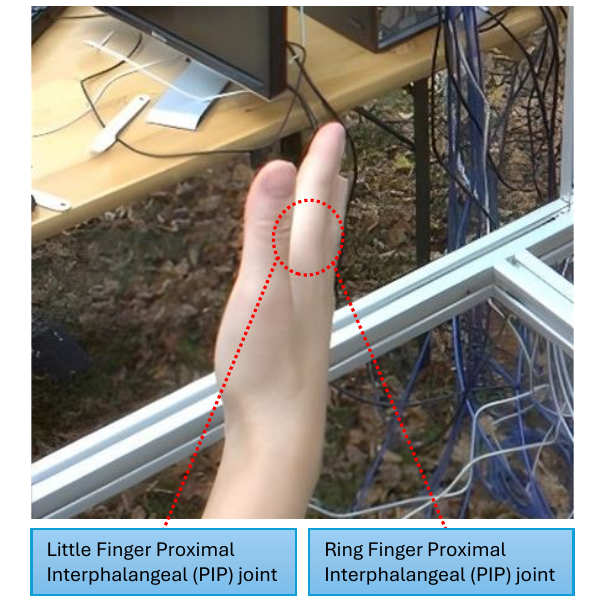}
\caption{Possible location of the `aligned' Little Finger Proximal Interphalangeal (PIP) joint and Ring Finger Proximal Interphalangeal (PIP) joint underneath the index and middle finger. The relationship along the x-axis for the two PIP joints is ambiguous, making it necessary to drop the relative position X information of the two joints.}
\label{fig:aligned_case}
\end{figure}

%% file: supp_sec/dataset/word_cloud.tex
\begin{figure*}[t]
\centering
\includegraphics[width=\textwidth]{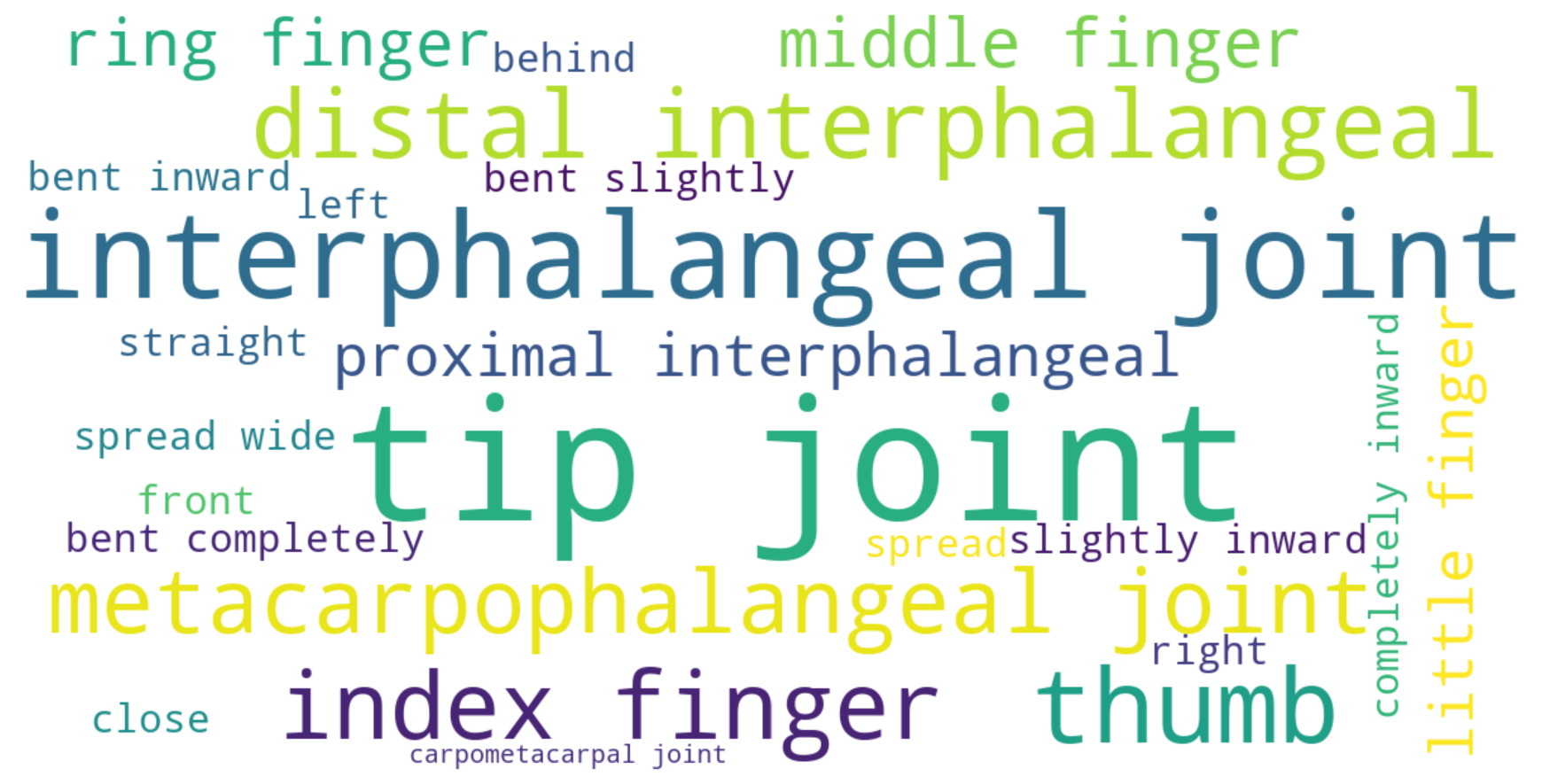}
\caption{\textbf{Word cloud representation of the most frequently used terms in the caption options extracted from our dataset.} Prominent anatomical terms like ``tip joint'', ``interphalangeal joint'', and ``metacarpophalangeal joint'' highlight the fine-grained spatial and anatomical focus of the hand-centric question-answer pairs.}
\label{fig:word_cloud}
\end{figure*}

%% file: supp_sec/dataset/train_eval_stat.tex
\begin{figure*}[t]
\centering
\includegraphics[width=\linewidth]{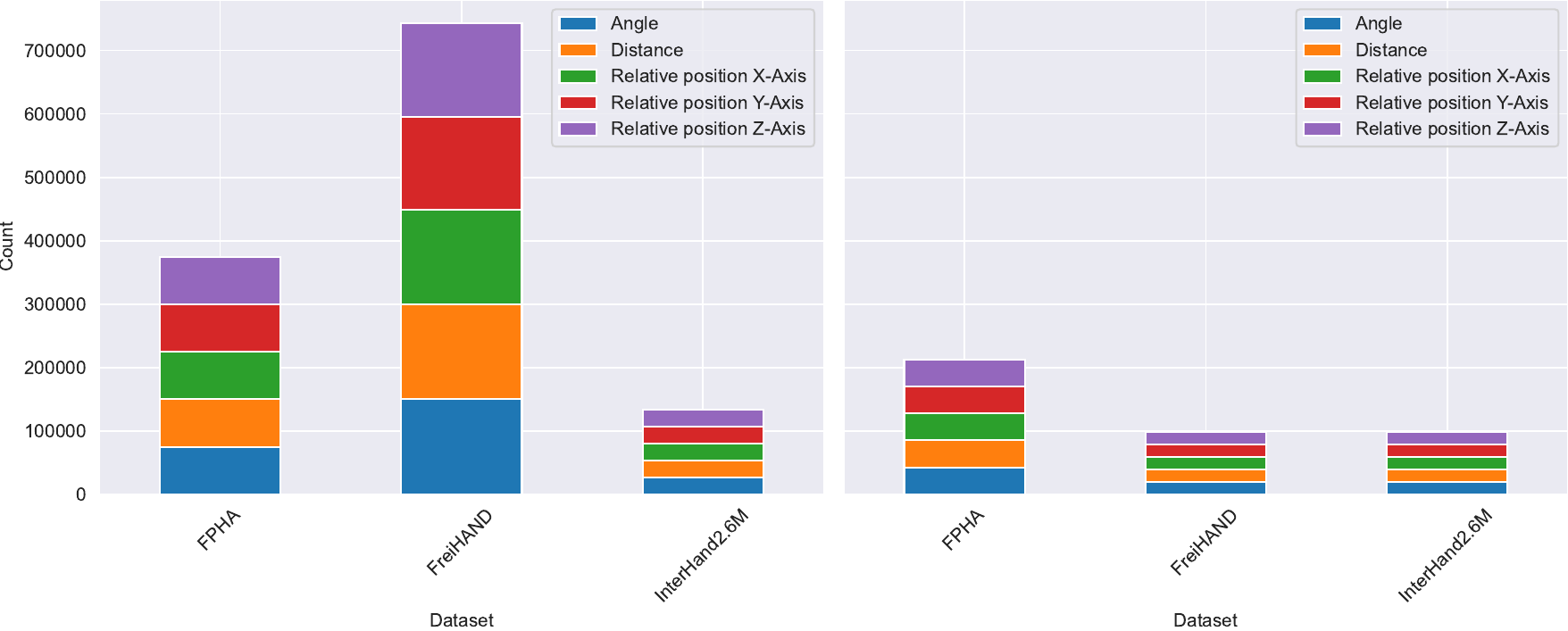}
\caption{\textbf{Breakdown of question types across the training (left) and evaluation (right) splits for each dataset in the HandVQA benchmark.} Each dataset contains a balanced distribution of all five spatial reasoning tasks: angle, distance, and relative positions along the X, Y, and Z axes. This uniformity supports fair evaluation across all pose subtasks.}
\label{fig:train_eval_stat}
\end{figure*}

%% file: supp_sec/dataset/catagory_stats.tex
\begin{table}[t]
\centering
\caption{Label frequency grouped by pose descriptors.}
\resizebox{\columnwidth}{!}{%
\begin{tabular}{ll}
\hline
\textbf{Pose Descriptors} & \multicolumn{1}{c}{\textbf{Label frequency (Count)}}                                                                                                 \\ \hline
Angle                     & \small {\begin{tabular}[c]{@{}l@{}}bent slightly inward (130,993), bent inward (107,143), \\ straight (74,007), bent completely inward (21,622)\end{tabular}} \\
Distance                  & \small {\begin{tabular}[c]{@{}l@{}}spread wide from (222,429), spread from (109,353), \\ close to (1,983)\end{tabular}}                                       \\
Rel. Pos. (X)             & \small {at the left of (198,988), at the right of (133,109)}                                                                                                  \\
Rel. Pos. (Y)             & \small {above (215,725), below (112,984)}                                                                                                                     \\
Rel. Pos. (Z)             & \small {in front of (177,216), behind (152,481)}                                                                                                              \\ \hline
\end{tabular}%
}
\label{tab:label_frequency_compact}
\end{table}

%% file: supp_sec/experiments.tex
\section{Training Details and Further Analysis on Experiments}
\label{sec:further_exp_analysis}

 This section provides a comprehensive overview of our training configuration alongside a series of additional analyses that expand upon the main experimental findings. We examine model behavior through confusion matrices, investigate cross-dataset transferability, study the effect of dataset-specific finetuning, compare VLM outputs with human judgments, and describe the construction of zero-shot evaluation datasets. 

\noindent \subsection{Training Setup}
\label{exp:training_setup}
We finetune all VLMs using LoRA\cite{hu2022lora} with rank 8 and alpha 32, targeting all linear layers. We use a learning rate of 1e-4. For all VLMs we train on FreiHAND VQA pairs for 1 epoch across 4 RTX 6000 ada GPUs and an Intel(R) Xeon(R) Gold 6326 CPU @ 2.90GHz CPU with a per device batch size 2, utilizing gradient accumulation over 16 steps for all datasets, resulting in an effective batch size of 128. We train InterHand2.6M VQA pairs for 3 epochs with a per device batch size of 1, resulting in an effective batch size of 64. In case of FPHA VQA pairs, we train for 1 epoch with a per device batch size of 1, resulting in an effective batch size of 64. \mub{All trainings are done on bfloat16 precision for speed, memory efficiency, and numerical stability.} We use the SWIFT~\cite{zhao2024swiftascalablelightweightinfrastructure} package to finetune all VLMs.

\subsection{Behavior of VLMs Across Spatial Descriptors}
\input{sec/figures/confusion_matrix/angle_conf}
\input{sec/figures/confusion_matrix/distance_conf}
\input{sec/figures/confusion_matrix/rel_x_conf}
\input{sec/figures/confusion_matrix/rel_y_conf}
\input{sec/figures/confusion_matrix/rel_z_conf}

Figures \ref{fig:angle_conf}, \ref{fig:angle_conf_ft}, \ref{fig:distance_conf_compare},
\ref{fig:rel_pos_x_conf_compare}, \ref{fig:rel_pos_y_conf_compare}, and \ref{fig:rel_pos_z_conf_compare} present confusion matrices for both base and fine-tuned VLMs, including DeepSeek \cite{lu2024deepseekvl}, LLaVA \cite{liu2023llava}, and Qwen-VL \cite{Qwen-VL}. The confusion matrices are constructed from the evaluation sets of all three datasets combined, enabling a direct comparison of model behavior before and after fine-tuning.

\subsubsection{Angle}

In Figure~\ref{fig:angle_conf}, we present the confusion matrix for the angle pose descriptor across four VLMs. A clear pattern emerges: all models exhibit a strong bias toward predicting the label ``bent slightly inward'', regardless of the actual ground truth. This bias dominates the prediction distribution across all ground truth categories.

In addition, each model shows a consistent preference ordering in its predictions across all ground truth classes. For instance, DeepSeek most frequently predicts ``bent slightly inward'', followed by ``bent inward'', then ``bent completely inward'', and finally ``straight''. This ordered bias persists even when the correct label is different. Similar trends are observed for LLaVA, and Qwen-VL, though the exact order of predicted preferences varies by model.

These results indicate that current VLMs lack the fine-grained spatial understanding required to accurately differentiate joint bending angles. Rather than interpreting the true angle from visual cues, models tend to default to mid-range or ambiguous options, revealing a limitation in their ability to reason about subtle variations in hand articulation.

After fine-tuning, this bias is substantially reduced as shown in Figure \ref{fig:angle_conf_ft}. The prediction distribution becomes more balanced, with a stronger concentration along the diagonal of the confusion matrix. This indicates improved discrimination of fine-grained joint angles. However, confusion between adjacent angle categories remains, suggesting that subtle variations in articulation are still challenging.

\subsubsection{Distance} 

In Figure~\ref{fig:distance_conf}, we present the confusion matrix for the distance pose descriptor, comparing model predictions across three distance-related spatial relationships: close to, spread from, and spread wide from.

In the distance pose descriptor, we observe a general bias toward predicting ``close to'' regardless of the ground truth distance label. This tendency is especially pronounced in LLaVA, and Qwen-VL, which frequently default to ``close to'' even when the actual relationship is ``spread from'' or ``spread wide from''. In contrast, DeepSeek demonstrates a more balanced prediction pattern across all three distance categories, indicating relatively better spatial discrimination.

While DeepSeek shows a slightly more distributed prediction pattern, it still tends to overpredict ``spread wide from''. The results suggest that VLMs struggle to distinguish varying levels of inter-joint distances from visual input alone.

This over-reliance on the ``close to'' class indicates that current models may not be effectively grounding physical separation between joints. Instead, they default to the most semantically neutral or "safe" spatial label when uncertain, mirroring the trends observed in the angle classification task. 

Following fine-tuning, the prediction distributions become significantly more balanced, with improved alignment along the diagonal (Figure \ref{fig:distance_conf_ft}). Models demonstrate better discrimination between different levels of inter-joint distance. Nevertheless, residual confusion persists between ``spread from'' and ``spread wide from'', indicating that fine-grained distance reasoning remains difficult.

\subsubsection{Relative Positions (X, Y and Z-axis)}

The confusion matrices for the relative position tasks along the X, Y, and Z axes (Figures~\ref{fig:rel_x_conf}, \ref{fig:rel_y_conf}, and \ref{fig:rel_z_conf}) show near uniform distributions with weak diagonal patterns for most of the cases, indicating behavior similar to random guessing—consistent with around 50\% accuracy observed across datasets in Table 3 of the main paper.

For the Y-axis, LLaVA overpredicts ``above'', while others show slightly more balanced but still unreliable outputs. The Z-axis shows the strongest bias: LLaVA and DeepSeek consistently overpredict ``in front of,'' failing to capture ``behind.''

In contrast, fine-tuning leads to a pronounced concentration along the diagonal across all axes, indicating substantial improvement in directional understanding (Figures \ref{fig:rel_x_conf_ft}, \ref{fig:rel_y_conf_ft}, and \ref{fig:rel_z_conf_ft}). The models learn to reliably distinguish binary spatial relations such as left–right, above–below, and front–behind. While minor confusion remains, particularly in depth (Z-axis), the overall spatial grounding is significantly enhanced.

\input{supp_sec/experiments/combine_results_angle_distance}
\input{supp_sec/experiments/combined_results_rel_pos}
\subsection{Further Experimental Comparisons}
We further investigate how finetuning strategies and dataset characteristics influence cross-dataset generalization and overall spatial reasoning performance of VLMs.

\subsubsection{Finetuned Models on Individual Datasets vs. Training on the Unified HandVQA Benchmark}

To better understand the impact of large-scale, multi-dataset supervision, we compare LLaVA Mistral 7B models finetuned on each dataset individually (FreiHAND, InterHand2.6M, FPHA) against a single model trained on the full HandVQA benchmark, which unifies all three datasets into a large and diverse supervision signal. We focus on LLaVA Mistral 7B because it was the strongest base model in our benchmark and exhibited the most stable finetuning behavior. Overall, the HandVQA-tuned model substantially outperforms the base model by a large margin across all datasets and all task types: Angle, Distance, and Relative Position reasoning (Tables \ref{tab:llava_mistral_handvqa_results} and \ref{tab:relpos_llava_mistral_handvqa}). This confirms that large-scale multimodal supervision with structured spatial queries transfers effectively and imparts generalizable spatial reasoning abilities that individual datasets alone cannot provide.

However, we also observe that the HandVQA-trained model does not always surpass the models finetuned on each dataset individually, especially in tasks where dataset specific characteristics dominate performance. For example, in FreiHAND and FPHA, the individually finetuned models achieve slightly higher accuracy or lower MAE on certain attributes. We attribute this phenomenon to the intrinsic limitations of LoRA finetuning. LoRA constrains parameter updates to a low rank decomposition, meaning that the model must compress all newly learned information across three large and heterogeneous datasets into a small number of trainable low rank matrices. As the diversity of training signals grows, these matrices must encode more variations, object contexts, hand configurations, and camera setups. This compression inevitably introduces information bottlenecks, leading to mild degradation on certain dataset specific details compared to models finetuned exclusively on a single, homogeneous dataset.
\subsubsection{Base Models vs. Cross-Dataset Performance}
\input{supp_sec/experiments/cross_dataset_angle_distance}
\input{supp_sec/experiments/cross_dataset_rel_pos}
\input{supp_sec/experiments/confidence_analysis/base/conf_analysis_base}
\input{supp_sec/experiments/confidence_analysis/finetuned/conf_analysis_finetuned}

To evaluate cross-dataset transfer, we compare base VLMs against models finetuned on the FreiHAND dataset and test them on two out-of-distribution datasets: InterHand2.6M (allocentric, multiview) and FPHA (egocentric, first-person). FreiHAND is predominantly allocentric, and therefore provides a controlled setup for studying how allocentric supervision transfers to both similar (allocentric) and different (egocentric) camera viewpoints.

As shown in Tables \ref{tab:ood_freihand_only} and \ref{tab:ood_freihand_relpos} across all descriptors Angle, Distance, and Relative Position FreiHAND finetuned models show substantial improvement when evaluated on InterHand2.6M. This trend is consistent across all three architectures (DeepSeek Janus Pro 7B, LLaVA Mistral 7B, and Qwen2.5 VL 7B Instr.), confirming that allocentric-to-allocentric transfer is highly effective. The improved performance suggests that the spatial reasoning learned from FreiHAND generalizes well to another multiview, third-person dataset that shares similar camera geometry and viewpoint distribution. However, the same finetuned models show less consistent gains when evaluated on the egocentric FPHA dataset. While Distance and Relative Position Y exhibit large improvements—likely because these descriptors depend more on coarse spatial relations rather than precise articulation—other attributes such as Angle and Relative Position Z remain challenging. These findings indicate that allocentric training alone does not fully prepare the model for the viewpoint distortions and hand-camera proximities inherent to egocentric perspectives.

\subsubsection{Comparison with Pure Vision Models}
\input{supp_sec/experiments/vision_only_baselines}
\camready{To assess the extent to which explicit multimodal supervision contributes to fine-grained spatial reasoning, we compare vision-language models against a strong pure vision baseline, HaMeR~\cite{pavlakos2024reconstructing}. HaMeR directly predicts 3D hand meshes from images, using our pipeline that was later converted into text description and evaluated against ground truth. The comparison is reported in Table~\ref{tab:vision_only}, covering Angle and Distance prediction tasks on the FreiHAND and InterHand2.6M test sets. The FPHA dataset is excluded, as it is not part of HaMeR’s training data.}

\camready{We first observe that the base LLaVA model performs significantly worse than HaMeR across all settings. For instance, on FreiHAND, LLaVA achieves only 42.48\% accuracy for Angle and 13.18\% for Distance, compared to 59.53\% and 88.86\% respectively, for HaMeR. A similar trend holds on InterHand2.6M. This gap highlights that, without task-specific supervision, VLMs lack the ability to extract precise geometric cues required for accurate hand pose reasoning.}

\camready{After finetuning on the HandVQA benchmark, LLaVA shows substantial improvements across all metrics, outperforming HaMeR on several tasks. In particular, LLaVA~\textit{ft} achieves higher Angle accuracy on FreiHAND (64.35\% vs. 59.53\%) and significantly improves Distance prediction on InterHand2.6M (90.27\% vs. 88.11\%). These results demonstrate that structured multimodal supervision can compensate for the lack of explicit geometric inductive bias in VLMs and enable them to learn fine-grained spatial relationships directly from data.}

\subsubsection{Model Confidence and Uncertainty Analysis}
\camready{We further analyze the reliability of model predictions by studying the relationship between predicted confidence and empirical accuracy.}

\camready{We initially explored estimating confidence via an additional classification head over discrete angle bins, jointly trained with the VLM. However, we observed inconsistent behavior between the auxiliary head outputs and the textual predictions, indicating a lack of alignment between internal representations and generated answers. Due to this inconsistency, we instead adopt a prompting-based approach for confidence estimation, following prior work on verbalized uncertainty in VLMs~\cite{groot-valdenegro-toro-2024-overconfidence}.}

\camready{Specifically, we prompt the LLaVA Mistral 7B to output likelihoods over discrete answer categories and evaluate calibration using reliability diagrams and confidence density histograms. This setup enables direct comparison between predicted confidence and empirical accuracy.}

\camready{As shown in Fig.~\ref{fig:base_model_pred_dist}, the base model exhibits a skewed confidence distribution, where a significant portion of incorrect predictions are associated with high confidence. This is further confirmed in the reliability diagram (Fig.~\ref{fig:base_reliability}), where predictions deviate substantially from the diagonal, indicating strong overconfidence. In other words, the model assigns high confidence even when accuracy is low.}

\camready{After finetuning, the prediction distribution shifts noticeably (Fig.~\ref{fig:finetuned_model_pred_dist}). While incorrect high-confidence predictions are reduced, the overall confidence mass moves toward lower and mid-range values. The corresponding reliability diagram (Fig.~\ref{fig:finetuned_reliability}) shows that confidence aligns more closely with accuracy, indicating improved calibration compared to the base model.}

\camready{However, a distinct behavior emerges: despite higher accuracy, the finetuned model exhibits systematically lower confidence. As observed in Fig.~\ref{fig:finetuned_reliability}, most points lie below the diagonal, suggesting a tendency toward underconfidence. That is, even correct predictions are often assigned conservative confidence scores.}

\camready{Overall, these results indicate that base VLMs suffer from overconfident errors, while finetuning reduces such failures but introduces underconfidence. This highlights that improved accuracy does not directly translate to well-calibrated uncertainty, and that reliable confidence estimation remains an open challenge even after task-specific supervision.}

\subsubsection{Failure Mode Analysis}
\input{supp_sec/experiments/easy_vs_hard}
\input{supp_sec/experiments/easy_vs_hard_vis}
We additionally analyze failure cases of the finetuned model by separating samples into \textit{Easy} and \textit{Hard} subsets based on ambiguity factors such as occlusion and interaction complexity. Quantitative results are summarized in Table~\ref{tab:hard_easy}, and representative qualitative examples are shown in Fig. \ref{fig:easy_vs_hard}.

As shown in Table~\ref{tab:hard_easy}, LLaVA Mistral 7B \textit{finetuned} achieves strong performance on Easy samples across all descriptors (e.g., 75.92\% for Angle and 95.01\% for Distance), but exhibits a consistent drop on Hard samples (55.1\% for Angle and 78.01\% for Distance). Similar degradation is observed across all relative position axes. This gap indicates that performance is primarily affected by visual ambiguity rather than uniformly limited across all samples.

Fig.~\ref{fig:easy_vs_hard} provides qualitative examples illustrating this behavior. In Easy cases (top row), the model correctly predicts both single-hand and hand-object interactions, where the spatial configuration is clearly visible and minimally occluded. In contrast, Hard cases (bottom row) show failure examples where occlusion, viewpoint, or interaction complexity makes the spatial relationship ambiguous, leading to incorrect predictions.

\subsection{Human Evaluation}
\input{supp_sec/experiments/human_eval}
To understand how current VLMs compare to human spatial reasoning, we conducted a small scale human study on a subset of HandVQA. As shown in Table \ref{tab:human_vs_vlm}, humans achieve 80.94\% accuracy, significantly outperforming all evaluated VLMs, which score between 41–46\%. This substantial gap highlights the difficulty of fine-grained hand pose reasoning, even for the strongest models such as LLaVA Mistral 7B and DeepSeek Janus Pro 7B. While VLMs can interpret coarse spatial relations, they frequently struggle with subtle joint-level distinctions such as small angular differences or depth ordering that humans can reliably discern. These results underscore the importance of specialized datasets like HandVQA for pushing VLMs toward human level performance in fine-grained 3D hand understanding.
\subsection{Zero-shot Evaluation Dataset Construction}
For our zero-shot evaluation, we consider two tasks: image-based gesture recognition and temporal sequence-based hand–object interaction recognition. The gesture recognition task is built from the HaGRID dataset \cite{Kapitanov_2024_WACV}, while the interaction recognition task uses the H2O dataset \cite{Kwon_2021_ICCV}. 

For the H2O interaction task, we directly use the action annotations provided in the dataset and convert them into multiple-choice questions (MCQs), each containing one correct answer and three distractors. In contrast, HaGRID provides only single-word gesture labels, which are insufficiently descriptive for zero-shot evaluation. To address this, we expand each gesture label into two semantically rich natural-language descriptions using Gemini \cite{gemini2024}, as illustrated in Table~\ref{tab:hagrid_descriptions}. These expanded descriptions provide the linguistic diversity necessary for evaluating zero-shot gesture reasoning. After generating gesture descriptions, we construct MCQs for each image, again with one correct description and three incorrect alternatives. To avoid semantic ambiguity, we group visually and semantically similar gestures such as \texttt{two\_up} and \texttt{two\_up\_inverted} into the same category. Consequently, when one appears as the correct option, the other is never used as an incorrect distractor. This ensures that evaluation difficulty arises from genuine reasoning challenges rather than annotation artifacts or label confusion.

\section{Qualitative Results}
Figure~\ref{fig:qualitative_gesture} and Figure~\ref{fig:qualitative_interaction} present qualitative zero-shot results on the HaGRID and H2O datasets, respectively. In the HaGRID gesture recognition task, the fine-tuned model consistently selects the correct semantic description, despite never having been trained on gesture sentences or gesture-specific supervision. The base model, by contrast, frequently misidentifies even visually obvious gestures. A similar trend appears in the H2O interaction recognition task: when reasoning over short temporal sequences, the fine-tuned model accurately identifies subtle object manipulations while the base model often predicts incorrect actions, demonstrating that joint-level training on HandVQA yields transferable spatial reasoning even in multi-frame, object-centric scenarios.

Figure~\ref{fig:qualitative_freihand}, Figure~\ref{fig:qualitative_interhand}, and Figure~\ref{fig:qualitative_fpha} provide qualitative comparisons on FreiHAND, InterHand2.6M, and FPHA. Each figure shows HandVQA-style MCQs spanning all five pose descriptors, together with predictions from both the base LLaVA model and its HandVQA-fine-tuned counterpart. Across all datasets, the base model most of the time chooses incorrect answers, including in cases involving clear geometry or simple articulation patterns. The fine-tuned model, however, consistently resolves the correct spatial relation, highlighting its ability to interpret fine-grained joint positions, bending angles, distances, and relative spatial orientations.

Finally, Figure~\ref{fig:qualitative_ood} examines generalization to in-the-wild images paired with questions phrased differently from the HandVQA templates and targeting higher-level finger-level geometry. The fine-tuned model reliably interprets these queries while the base model fails. These examples demonstrate that fine-grained joint-level supervision not only improves in-domain performance but also enables robust transfer to higher-level geometric reasoning.

\label{sec:qualitative_res}
\input{supp_sec/license_details}

\input{supp_sec/gesture_annotations}
\input{supp_sec/figures/qualitative_results_gesture}
\input{supp_sec/figures/qualitative_results_interaction}
\input{supp_sec/figures/qualitative_results_freihand}
\input{supp_sec/figures/qualitative_results_interhand}
\input{supp_sec/figures/qualitative_results_fpha}

\input{supp_sec/figures/qualitative_results_ood}

%% file: sec/figures/confusion_matrix/angle_conf.tex
\begin{figure*}[t!]
\centering
\includegraphics[width=\linewidth]{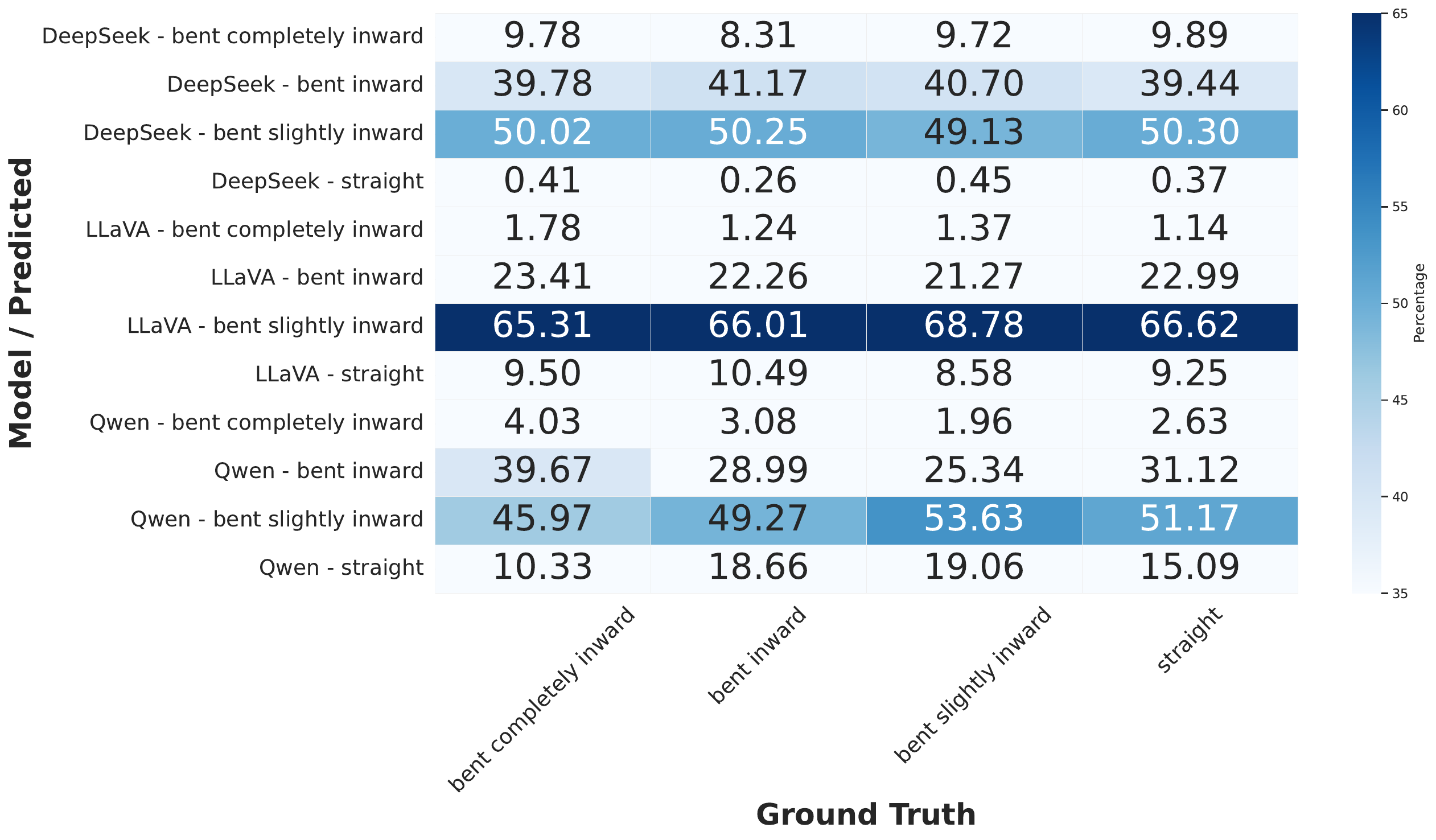}
\caption{\textbf{Angle confusion matrix across three VLMs.} All models frequently predict \textit{“bent slightly inward”} regardless of ground truth, revealing a strong prediction bias. Each model also follows a consistent preference ordering in its outputs, indicating difficulty in distinguishing fine-grained joint angles.}
\label{fig:angle_conf}
\end{figure*}

\begin{figure*}[t]
\centering
\includegraphics[width=\linewidth]{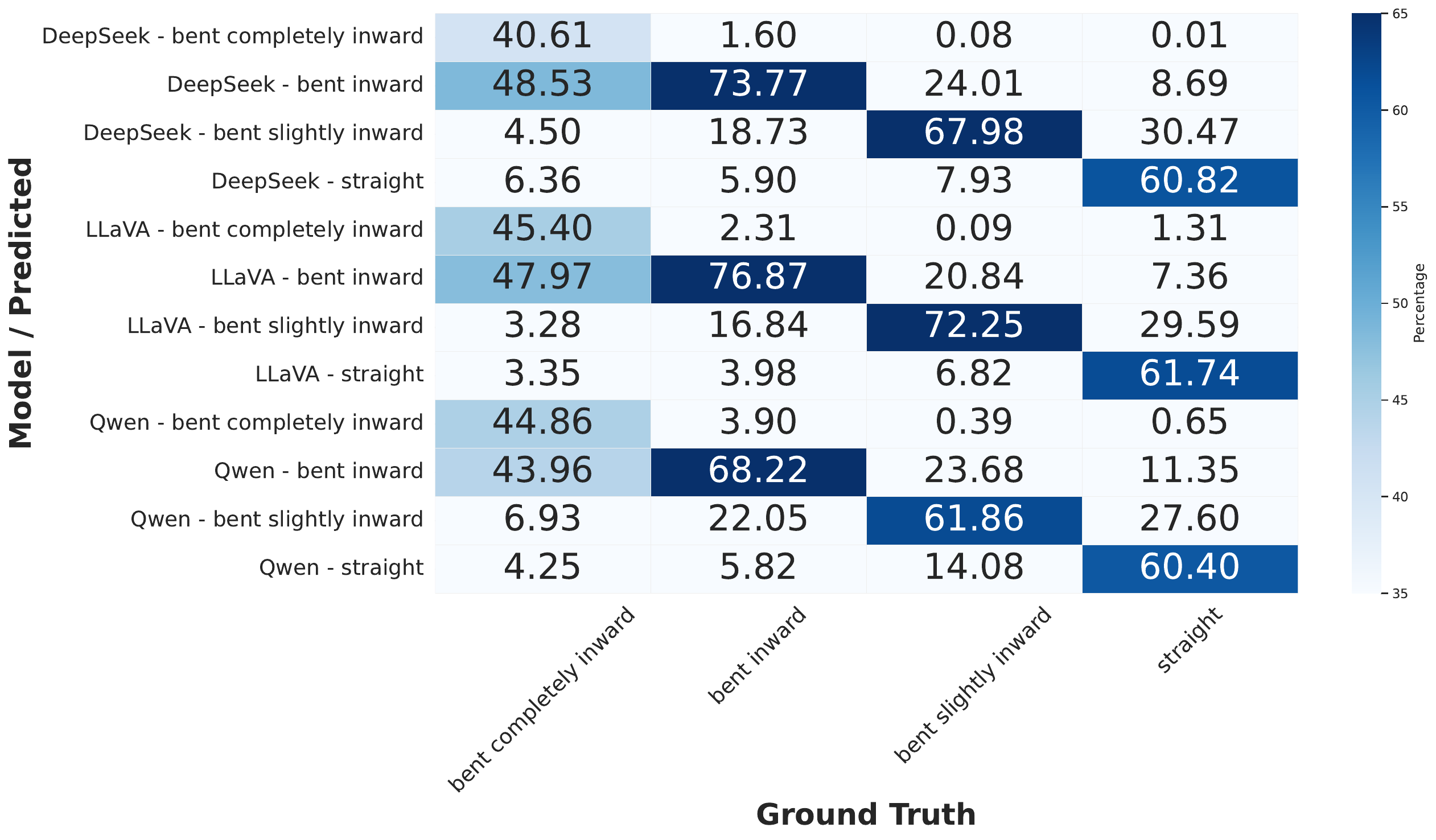}
\caption{\textbf{Angle confusion matrix across three fine-tuned VLMs.} Fine-tuning significantly reduces the dominant bias toward \textit{“bent slightly inward”} observed in the base models, resulting in stronger alignment along the diagonal. The models exhibit improved discrimination across angle categories, particularly for \textit{“bent inward”} and \textit{“straight”}, although some residual confusion remains between adjacent angle classes, indicating that fine-grained distinctions are still challenging.}
\label{fig:angle_conf_ft}
\end{figure*}

%% file: sec/figures/confusion_matrix/distance_conf.tex


\begin{figure*}[t!]
    \centering
    \begin{subfigure}[t]{0.49\linewidth}
        \centering
        \includegraphics[width=\linewidth]{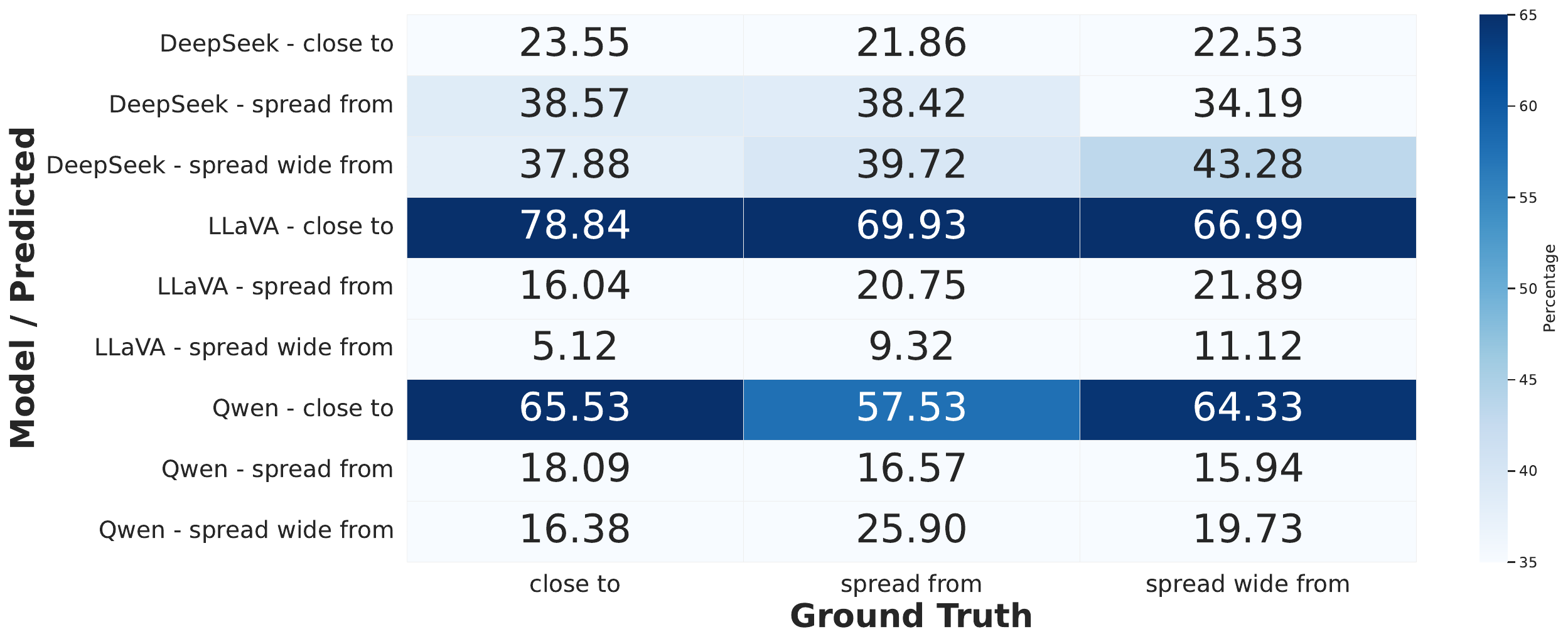}
        \caption{Base Models.}
        
        \label{fig:distance_conf}
    \end{subfigure}
    \hfill
    \begin{subfigure}[t]{0.49\linewidth}
        \includegraphics[width=\linewidth]{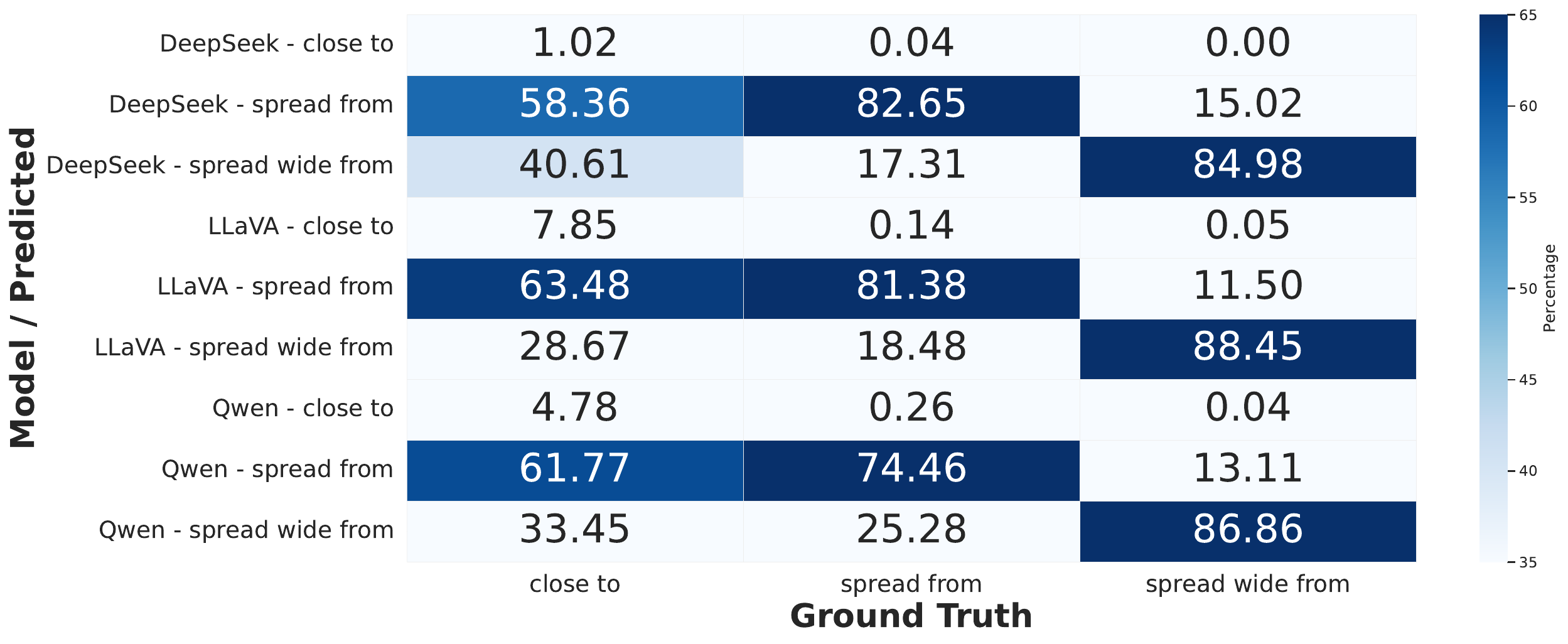}
        \caption{Finetuned Models.}
        
        \label{fig:distance_conf_ft}
    \end{subfigure}
    \caption{\textbf{Distance confusion matrix comparison between base and fine-tuned VLMs.} The base models exhibit a strong bias toward predicting \textit{“close to”} across multiple ground-truth categories, leading to a skewed prediction distribution. In contrast, fine-tuning produces a more balanced distribution with increased alignment along the diagonal, indicating improved discrimination between distance categories. However, residual confusion persists between \textit{“spread from”} and \textit{“spread wide from”}, suggesting continued difficulty in distinguishing fine-grained spatial separations.}
    \label{fig:distance_conf_compare}
\end{figure*}

%% file: sec/figures/confusion_matrix/rel_x_conf.tex

\begin{figure*}[t!]
    \centering
    \begin{subfigure}[t]{0.49\linewidth}
        \centering
        \includegraphics[width=\linewidth]{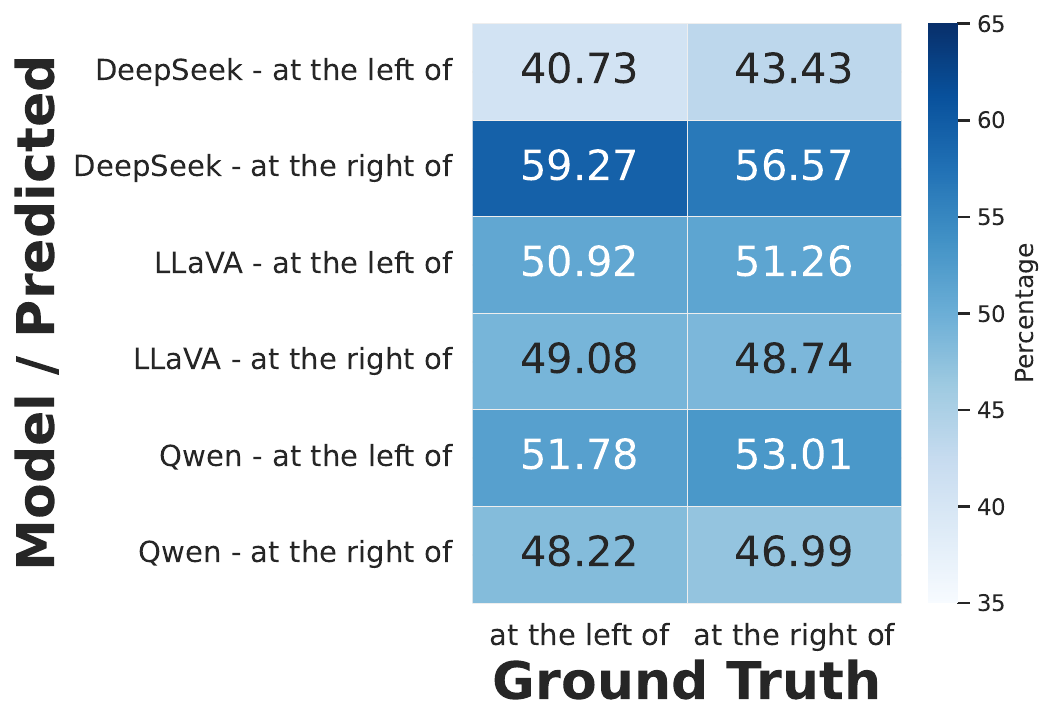}
        \caption{Base models.}
        \label{fig:rel_x_conf}
    \end{subfigure}
    \hfill
    \begin{subfigure}[t]{0.49\linewidth}
        \centering
        \includegraphics[width=\linewidth]{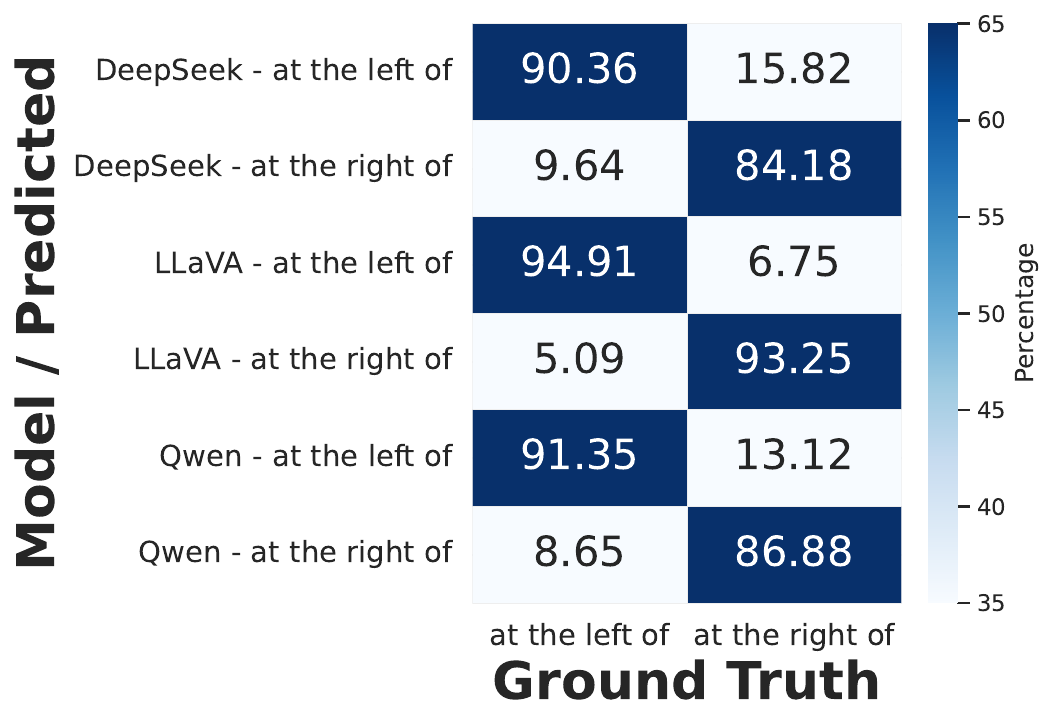}
        \caption{Finetuned models.}
        \label{fig:rel_x_conf_ft}
    \end{subfigure}
    \caption{\textbf{Relative Position X confusion matrix comparison between base and fine-tuned VLMs.} The base models exhibit near-random predictions between \textit{“at the left of”} and \textit{“at the right of”}, indicating a lack of reliable directional understanding. In contrast, fine-tuning leads to a strong concentration along the diagonal, demonstrating significantly improved left–right discrimination. This suggests that relative positional reasoning along the horizontal axis is effectively learned after fine-tuning, with minimal residual confusion.}
    \label{fig:rel_pos_x_conf_compare}
\end{figure*}

%% file: sec/figures/confusion_matrix/rel_y_conf.tex

\begin{figure*}[t!]
    \centering
    \begin{subfigure}[t]{0.49\linewidth}
        \centering
        \includegraphics[width=\linewidth]{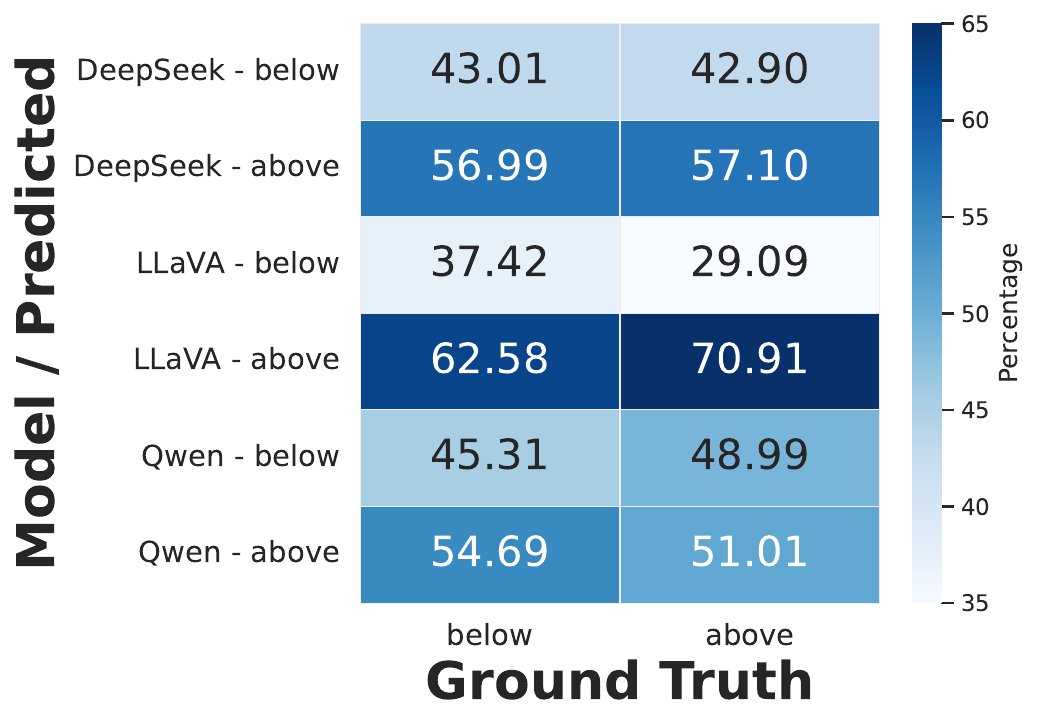}
        \caption{Base models.}
        \label{fig:rel_y_conf}
    \end{subfigure}
    \hfill
    \begin{subfigure}[t]{0.49\linewidth}
        \centering
        \includegraphics[width=\linewidth]{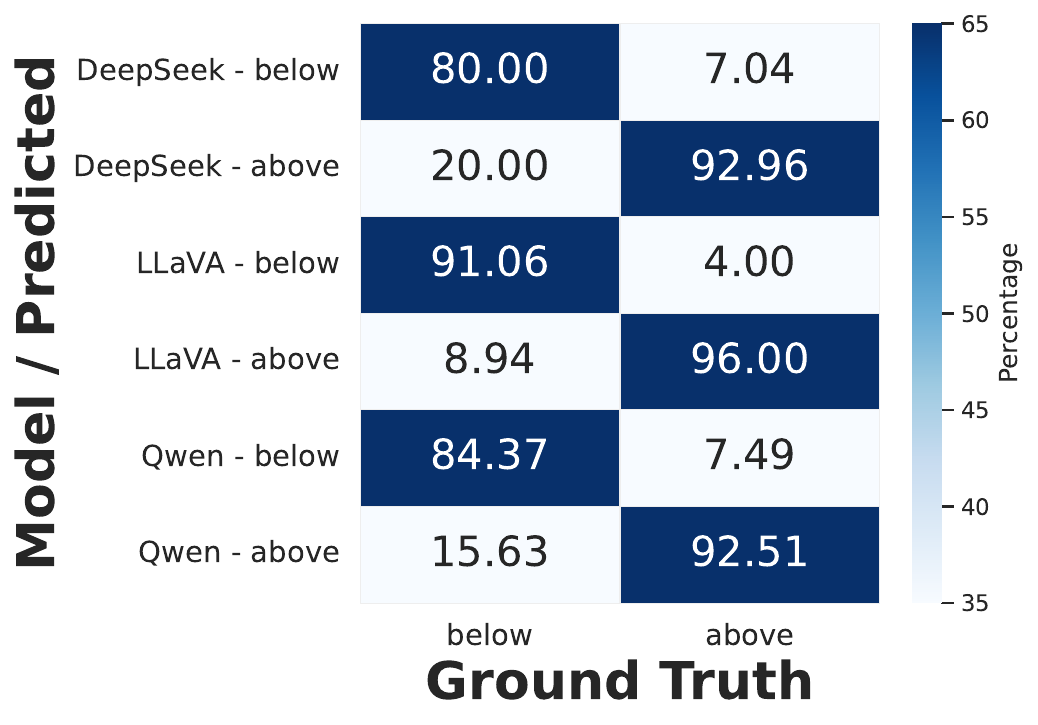}
        \caption{Finetuned models.}
        \label{fig:rel_y_conf_ft}
    \end{subfigure}
    \caption{\textbf{Relative Position Y confusion matrix comparison between base and fine-tuned VLMs.} The base models exhibit inconsistent predictions between \textit{“below”} and \textit{“above”}, reflecting weak vertical positional understanding. In contrast, fine-tuning results in a strong alignment along the diagonal, indicating substantially improved discrimination of vertical relationships. While minor confusion remains, the fine-tuned models demonstrate a clear and consistent understanding of relative position along the vertical axis.}
    \label{fig:rel_pos_y_conf_compare}
\end{figure*}

%% file: sec/figures/confusion_matrix/rel_z_conf.tex

\begin{figure*}[t!]
    \centering
    \begin{subfigure}[t]{0.49\linewidth}
        \centering
        \includegraphics[width=\linewidth]{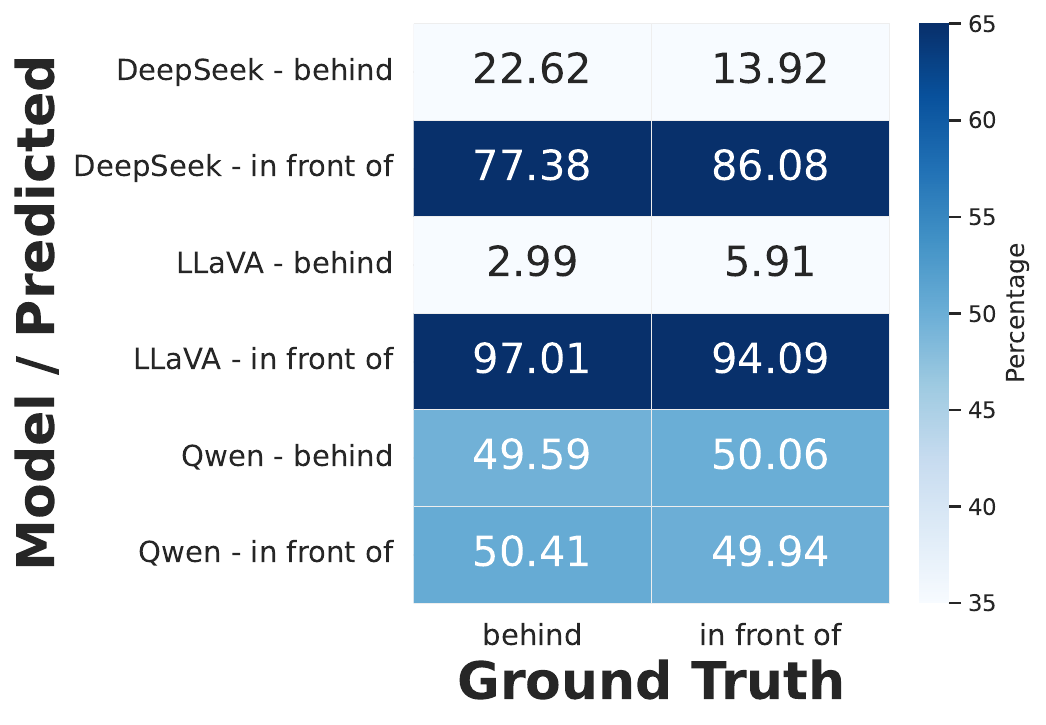}
        \caption{Base models.}
        \label{fig:rel_z_conf}
    \end{subfigure}
    \hfill
    \begin{subfigure}[t]{0.49\linewidth}
        \centering
        \includegraphics[width=\linewidth]{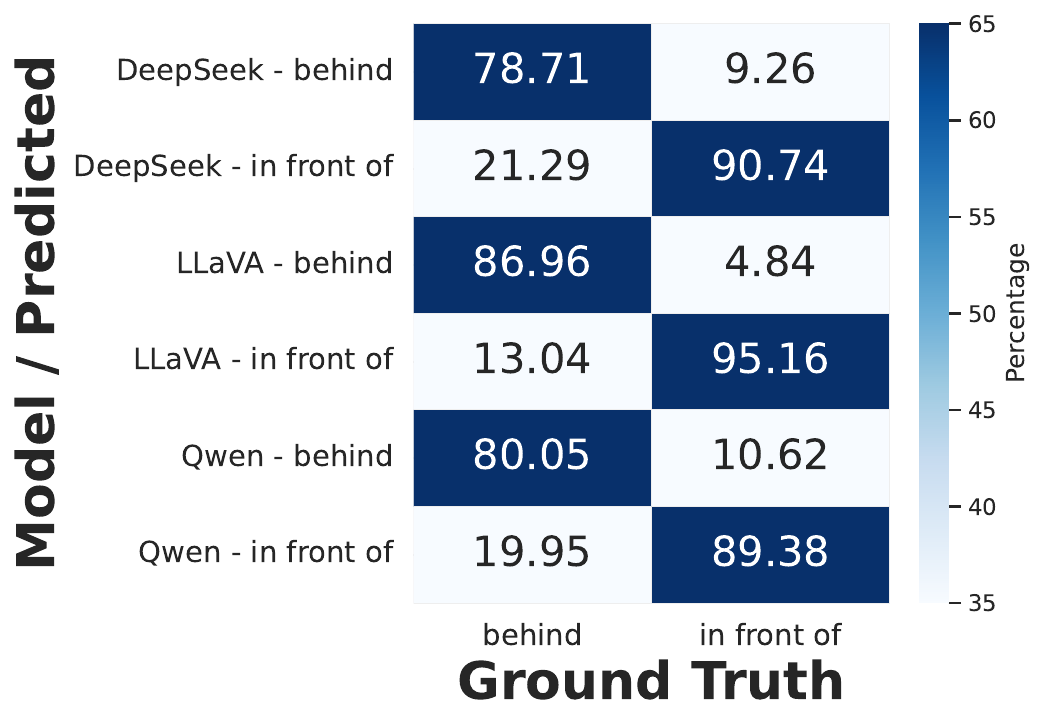}
        \caption{Finetuned models.}
        \label{fig:rel_z_conf_ft}
    \end{subfigure}
    \caption{\textbf{Relative Position Z confusion matrix comparison between base and fine-tuned VLMs.} The base models exhibit strong bias toward predicting \textit{“in front of”}, leading to highly skewed outputs and poor discrimination of depth relationships. In contrast, fine-tuning significantly improves alignment with the diagonal, indicating enhanced understanding of front–back spatial relations. While minor confusion persists, the fine-tuned models demonstrate substantially improved depth-aware reasoning compared to the base models.}
    \label{fig:rel_pos_z_conf_compare}
\end{figure*}

%% file: supp_sec/experiments/combine_results_angle_distance.tex
\scriptsize
\setlength{\tabcolsep}{0.5pt}          
\renewcommand{\arraystretch}{0.95}      

\begin{table*}[!t]
\caption{Angle and Distance results of LLaVA Mistral 7B finetuned models on Individual Datasets vs. finetuned on the Unified HandVQA Benchmark.}
  \centering
  \small

  \begin{tabular*}{\textwidth}{%
      @{\extracolsep{\fill}}
      lll
      @{\hspace{4pt}}
      crrr
      @{\extracolsep{\fill}}
  }
    \toprule
    \multirow{2}{*}{Model} & \multirow{2}{*}{Tuned} & \multirow{2}{*}{Eval}
      & \multicolumn{2}{c}{Angle} & \multicolumn{2}{c}{Distance} \\
    \cmidrule(lr){4-5} \cmidrule(lr){6-7}
        &   &   & Accuracy $\uparrow$ & MAE $\downarrow$ & Accuracy $\uparrow$ & MAE $\downarrow$ \\
    \midrule \addlinespace
    \multicolumn{7}{@{}l}{\textbf{Finetuned Models on Individual Datasets}}\\
    LLaVA Mistral 7B & InterHand2.6M & InterHand2.6M & 74.35 & 0.263 & 90.79 & 0.094 \\
    LLaVA Mistral 7B & FreiHAND  & FreiHAND  & 62.91 & 0.382 & 86.19 & 0.141 \\
    LLaVA Mistral 7B & FPHA      & FPHA      & 68.37 & 0.401 & 83.99 & 0.161 \\

    \addlinespace
    \multicolumn{7}{@{}l}{\textbf{Finetuned Model on HandVQA}}\\
    LLaVA Mistral 7B & HandVQA & InterHand2.6M & 72.26 & 0.283 & 90.27 & 0.099 \\
    LLaVA Mistral 7B & HandVQA & FreiHAND      & 64.35 & 0.367 & 86.71 & 0.136 \\
    LLaVA Mistral 7B & HandVQA & FPHA          & 67.95 & 0.411 & 84.64 & 0.154 \\

    \midrule
  \end{tabular*}

  \label{tab:llava_mistral_handvqa_results}
\end{table*}
\normalsize

%% file: supp_sec/experiments/combined_results_rel_pos.tex
\scriptsize
\setlength{\tabcolsep}{18pt}  
\renewcommand{\arraystretch}{0.95}

\begin{table*}[!t]
\caption{Relative Position accuracies of LLaVA Mistral 7B finetuned models on Individual Datasets vs. finetuned on the Unified HandVQA Benchmark.}
  \centering
  \small

  \begin{tabular*}{\textwidth}{@{}lll@{\hspace{4pt}}@{\extracolsep{\stretch{1}}}rrr@{}}
    \toprule
    \multirow{2}{*}{Model} & \multirow{2}{*}{Tuned} & \multirow{2}{*}{Eval}
      & \multicolumn{1}{c}{Rel.\ Pos.\ X}
      & \multicolumn{1}{c}{Rel.\ Pos.\ Y}
      & \multicolumn{1}{c}{Rel.\ Pos.\ Z} \\
    \cmidrule(lr){4-4} \cmidrule(lr){5-5} \cmidrule(lr){6-6}
        &   &   & Accuracy $\uparrow$ & Accuracy $\uparrow$ & Accuracy $\uparrow$ \\
    \midrule \addlinespace

    \multicolumn{6}{@{}l}{\textbf{Finetuned Models on Individual Datasets}}\\
    LLaVA Mistral 7B & InterHand2.6M & InterHand2.6M & 97.14 & 98.77 & 96.82 \\
    LLaVA Mistral 7B & FreiHAND  & FreiHAND  & 92.60 & 93.20 & 88.17 \\
    LLaVA Mistral 7B & FPHA      & FPHA      & 93.81 & 92.80 & 90.25 \\

    \addlinespace
    \multicolumn{6}{@{}l}{\textbf{Finetuned Model on HandVQA}}\\

    LLaVA Mistral 7B & HandVQA & InterHand2.6M & 96.15 & 98.66 & 96.74 \\
    LLaVA Mistral 7B & HandVQA & FreiHAND      & 93.82 & 94.34 & 90.24 \\
    LLaVA Mistral 7B & HandVQA & FPHA          & 95.12 & 93.26 & 90.30 \\

    \midrule
  \end{tabular*}

  \label{tab:relpos_llava_mistral_handvqa}
\end{table*}
\normalsize

%% file: supp_sec/experiments/cross_dataset_angle_distance.tex
\scriptsize
\setlength{\tabcolsep}{0.5pt}          
\renewcommand{\arraystretch}{0.95}

\begin{table*}[!t]
\caption{Angle and Distance results of Base Models vs Out-of-Distribution Finetuned Models.}
  \centering
  \small
  \begin{tabular*}{\textwidth}{%
      @{\extracolsep{\fill}}
      lll
      @{\hspace{4pt}}
      rrrr
      @{\extracolsep{\fill}}
  }
    \toprule
    \multirow{2}{*}{Model} & \multirow{2}{*}{Tuned} & \multirow{2}{*}{Eval}
      & \multicolumn{2}{c}{Angle} & \multicolumn{2}{c}{Distance} \\
    \cmidrule(lr){4-5} \cmidrule(lr){6-7}
        &   &   & Accuracy $\uparrow$ & MAE $\downarrow$ & Accuracy $\uparrow$ & MAE $\downarrow$ \\
    \midrule \addlinespace
    \multicolumn{7}{@{}l}{\textbf{Base model (no tuning)}}\\
    DeepSeek Janus Pro 7B & – & InterHand2.6M & 34.10 & 0.883 & 45.55 & 0.657 \\
    DeepSeek Janus Pro 7B & – & FPHA      & 26.46 & 0.991 & 39.02 & 0.819 \\

    \addlinespace
    \multicolumn{7}{@{}l}{\textbf{Finetuned Models}}\\
    DeepSeek Janus Pro 7B & FreiHAND & InterHand2.6M & 56.15 & 0.456 & 82.70 & 0.175 \\
    DeepSeek Janus Pro 7B & FreiHAND & FPHA          & 25.52 & 1.028 & 79.73 & 0.203 \\

    \midrule \addlinespace
    \multicolumn{7}{@{}l}{\textbf{Base model (no tuning)}}\\
    LLaVA Mistral 7B & – & InterHand2.6M & 40.08 & 0.739 & 16.20 & 1.293 \\
    LLaVA Mistral 7B & – & FPHA      & 23.38 & 1.011 & 13.57 & 1.353 \\

    \addlinespace
    \multicolumn{7}{@{}l}{\textbf{Finetuned Models}}\\
    LLaVA Mistral 7B      & FreiHAND & InterHand2.6M & 56.81 & 0.447 & 83.26 & 0.170 \\
    LLaVA Mistral 7B      & FreiHAND & FPHA          & 26.45 & 1.015 & 79.94 & 0.201 \\

    \midrule \addlinespace
    \multicolumn{7}{@{}l}{\textbf{Base model (no tuning)}}\\
    Qwen 2.5 VL 7B Instr. & – & InterHand2.6M & 37.92 & 0.779 & 19.58 & 1.247 \\
    Qwen 2.5 VL 7B Instr. & – & FPHA      & 24.22 & 1.055 & 18.03 & 1.306 \\

    \addlinespace
    \multicolumn{7}{@{}l}{\textbf{Finetuned Models}}\\
    Qwen 2.5 VL 7B Instr. & FreiHAND & InterHand2.6M & 50.85 & 0.536 & 80.67 & 0.196 \\
    Qwen 2.5 VL 7B Instr. & FreiHAND & FPHA          & 24.65 & 1.083 & 78.94 & 0.211 \\

    \bottomrule
  \end{tabular*}
  \label{tab:ood_freihand_only}
\end{table*}
\normalsize

%% file: supp_sec/experiments/cross_dataset_rel_pos.tex
\scriptsize
\setlength{\tabcolsep}{5pt}         
\renewcommand{\arraystretch}{0.95}  

\begin{table*}[!t]
  \centering
  \small
  \caption{Relative-Position Results of Base Models vs Out-of-Distribution Finetuned Models.}
  \label{tab:ood_freihand_relpos}

  \begin{tabular*}{\textwidth}{@{}lll@{\hspace{4pt}}@{\extracolsep{\stretch{1}}}rrr@{}}
    \toprule
    \multirow{2}{*}{Model} & \multirow{2}{*}{Tuned} & \multirow{2}{*}{Eval}
      & \multicolumn{1}{c}{Rel.\ Pos.\ X}
      & \multicolumn{1}{c}{Rel.\ Pos.\ Y}
      & \multicolumn{1}{c}{Rel.\ Pos.\ Z} \\
    \cmidrule(lr){4-4} \cmidrule(lr){5-5} \cmidrule(lr){6-6}
        &   &   & Accuracy $\uparrow$ & Accuracy $\uparrow$ & Accuracy $\uparrow$ \\
    \midrule \addlinespace
    \multicolumn{6}{@{}l}{\textbf{Base model (no tuning)}}\\
    DeepSeek Janus Pro 7B & – & InterHand2.6M & 50.41 & 52.46 & 51.16 \\
    DeepSeek Janus Pro 7B & – & FPHA      & 43.02 & 52.64 & 61.73 \\

    \addlinespace
    \multicolumn{6}{@{}l}{\textbf{Finetuned Models}}\\

    DeepSeek Janus Pro 7B & FreiHAND & InterHand2.6M & 77.32 & 89.80 & 75.33 \\
    DeepSeek Janus Pro 7B & FreiHAND & FPHA          & 50.58 & 74.96 & 48.45 \\

    \midrule \addlinespace
    \multicolumn{6}{@{}l}{\textbf{Base model (no tuning)}}\\
    LLaVA Mistral 7B & – & InterHand2.6M & 49.72 & 66.26 & 40.87 \\
    LLaVA Mistral 7B & – & FPHA      & 50.27 & 56.33 & 56.73 \\

    \addlinespace
    \multicolumn{6}{@{}l}{\textbf{Finetuned Models}}\\
    LLaVA Mistral 7B      & FreiHAND & InterHand2.6M & 84.53 & 92.73 & 84.49 \\
    LLaVA Mistral 7B      & FreiHAND & FPHA          & 50.27 & 78.04 & 56.65 \\

    \midrule \addlinespace
    \multicolumn{6}{@{}l}{\textbf{Base model (no tuning)}}\\
    Qwen 2.5 VL 7B Instr. & – & InterHand2.6M & 48.98 & 49.78 & 49.33 \\
    Qwen 2.5 VL 7B Instr. & – & FPHA      & 50.98 & 48.53 & 49.79 \\

    \addlinespace
    \multicolumn{6}{@{}l}{\textbf{Finetuned Models}}\\
    Qwen 2.5 VL 7B Instr. & FreiHAND & InterHand2.6M & 77.61 & 86.27 & 66.98 \\
    Qwen 2.5 VL 7B Instr. & FreiHAND & FPHA          & 55.82 & 70.34 & 60.07 \\

    \bottomrule
  \end{tabular*}
\end{table*}
\normalsize

%% file: supp_sec/experiments/confidence_analysis/base/conf_analysis_base.tex
\begin{figure*}[t]
    \centering
    \begin{minipage}[t]{0.48\linewidth}
        \centering
        \includegraphics[width=\linewidth]{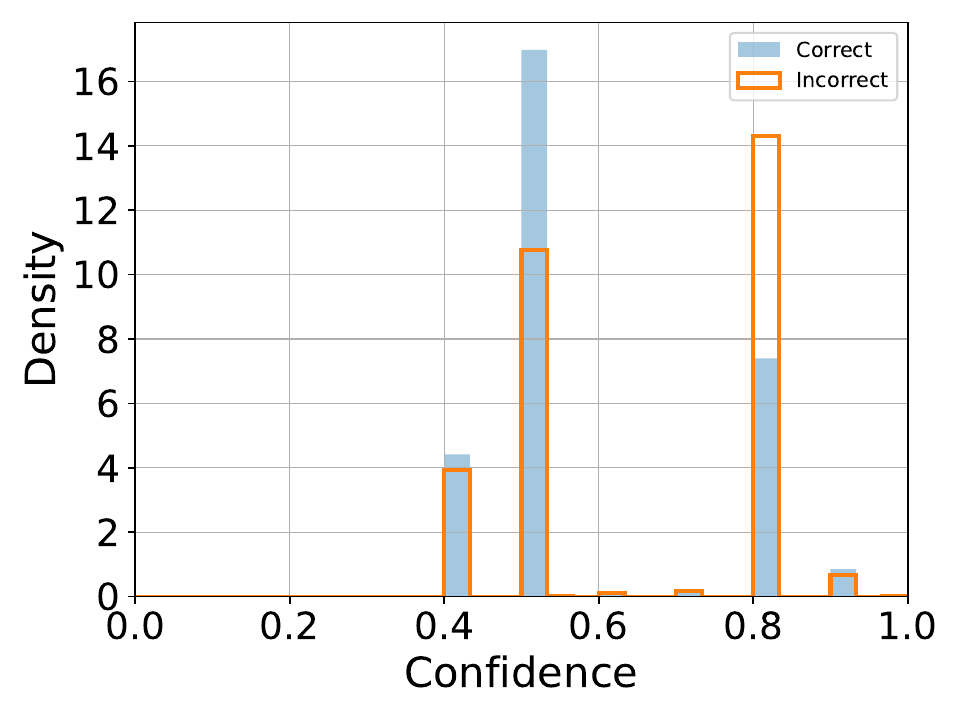}
        \caption{\textbf{Base Model Prediction Distribution (LLaVA-Mistral-7B).}
        Distribution of predicted confidence for correct and incorrect responses in the base LLaVA-Mistral-7B model. The confidence distribution is skewed, with a substantial portion of incorrect predictions assigned high confidence, revealing pronounced overconfident errors.}
        \label{fig:base_model_pred_dist}
    \end{minipage}
    \hfill
    \begin{minipage}[t]{0.48\linewidth}
        \centering
        \includegraphics[width=\linewidth]{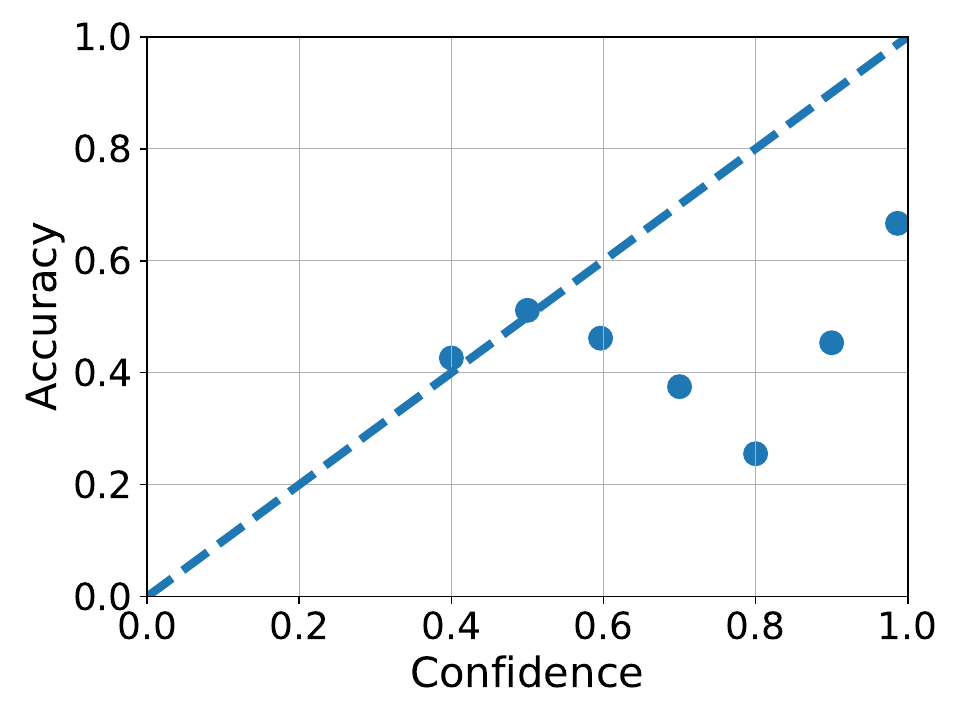}
        \caption{\textbf{Base Model Reliability Diagram (LLaVA-Mistral-7B).} Reliability diagram comparing predicted confidence and accuracy. The model is poorly calibrated and overconfident. Although accuracy is higher at confidence > 0.8, such predictions are rare (Fig. \ref{fig:base_model_pred_dist}), limiting their impact.}
        \label{fig:base_reliability}
    \end{minipage}
\end{figure*}

%% file: supp_sec/experiments/confidence_analysis/finetuned/conf_analysis_finetuned.tex
\begin{figure*}[t]
    \centering
    \begin{minipage}[t]{0.48\linewidth}
        \centering
        \includegraphics[width=\linewidth]{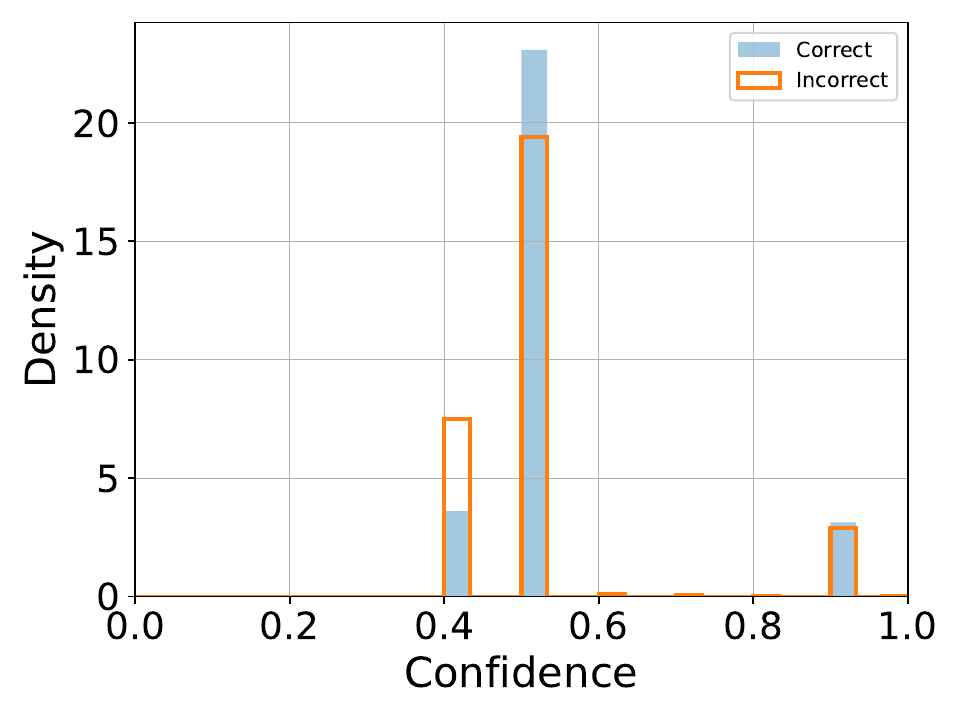}
        \caption{\textbf{Finetuned Model Prediction Distribution (LLaVA-Mistral-7B).}
        Distribution of predicted confidence for correct and incorrect responses after finetuning LLaVA-Mistral-7B. High-confidence incorrect predictions are reduced, and the confidence mass shifts toward lower and mid-range values, reflecting more conservative predictions.}
        \label{fig:finetuned_model_pred_dist}
    \end{minipage}
    \hfill
    \begin{minipage}[t]{0.48\linewidth}
        \centering
        \includegraphics[width=\linewidth]{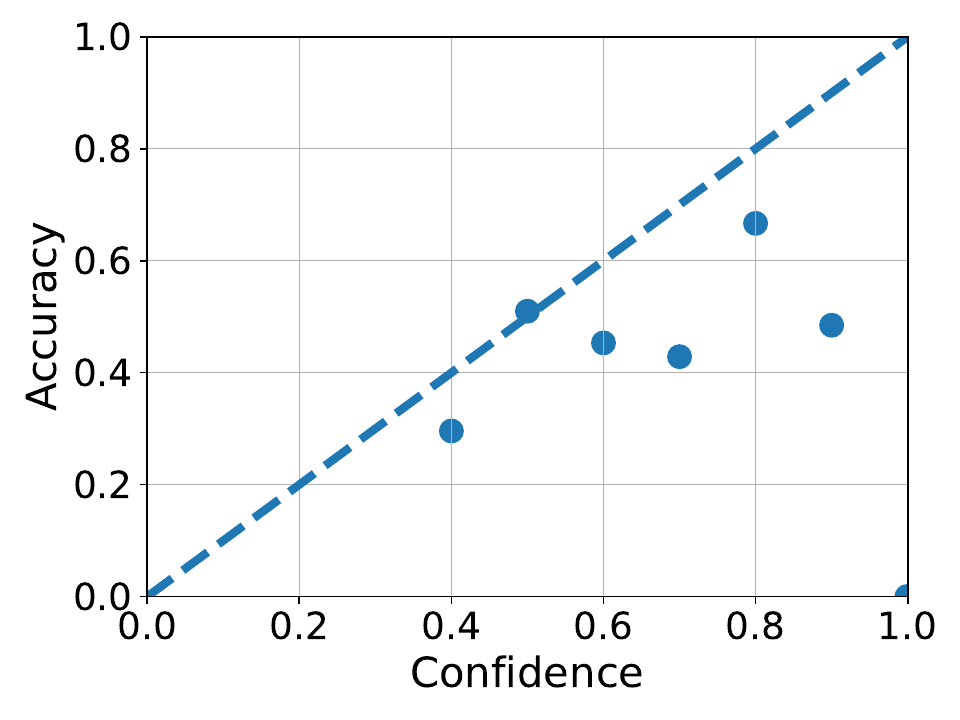}
        \caption{\textbf{Finetuned Model Reliability Diagram (LLaVA-Mistral-7B).}
        Reliability diagram of the finetuned LLaVA-Mistral-7B model. Confidence aligns more closely with empirical accuracy, indicating improved calibration. However, most points lie below the diagonal, suggesting systematic underconfidence despite higher accuracy.}
        \label{fig:finetuned_reliability}
    \end{minipage}
\end{figure*}

%% file: supp_sec/experiments/vision_only_baselines.tex
\begin{table*}[t]
    \centering
    \caption{Pure vision model (HaMeR \cite{pavlakos2024reconstructing}) vs. LLaVA Mistral 7B.}
    \setlength{\tabcolsep}{2pt} 
    \begin{tabular*}{\linewidth}{@{\extracolsep{\fill}}lcc@{\hspace{6pt}}||cc}
    \toprule
    \multirow{2}{*}{\textbf{Model}}
    & \multicolumn{2}{c||}{\textbf{FreiHAND test}}
    & \multicolumn{2}{c}{\textbf{InterHand2.6M test}} \\
    \cmidrule[\heavyrulewidth]{2-5}
    & \textbf{Angle} & \textbf{Distance} & \textbf{Angle} & \textbf{Distance} \\
    & Accuracy \% (MAE) & Accuracy \% (MAE) &  Accuracy \% (MAE) &  Accuracy \% (MAE) \\
    \midrule
    HaMeR & 59.53 (0.428) & \textbf{88.86 (0.113)} & \textbf{78.82 (0.218)} & 88.11 (0.119) \\
    LLaVA Mistral 7B & 42.48 (0.678) & 13.18 (1.342) & 40.08 (0.739) & 16.20 (1.293) \\
    LLaVA Mistral 7B \textit{finetuned} & \textbf{64.35 (0.367)} & 86.71 (0.136) & 72.26 (0.283) & \textbf{90.27 (0.099)} \\
    \bottomrule
\end{tabular*}
    \label{tab:vision_only}
\end{table*}

%% file: supp_sec/experiments/easy_vs_hard.tex
\begin{table*}[t!]
    \centering
    \setlength{\tabcolsep}{2pt} 
    \caption{ LLaVA Mistral 7B \textit{finetuned} on FreiHAND: Easy (225 QA), Hard (223 QA).}
    \begin{tabular*}{\linewidth}{@{\extracolsep{\fill}}lccccc}
        \toprule
        \textbf{Difficulty} & \textbf{Angle} & \textbf{Distance} & \textbf{R. Pos.(X)} & \textbf{R. Pos.(Y)} & \textbf{R. Pos.(Z)} \\
        & Accuracy \% (MAE) & Accuracy \% (MAE) & Accuracy \% & Accuracy \% & Accuracy \% \\
        \midrule
        Easy & \textbf{75.92 (0.255)} & \textbf{95.01 (0.08)} & \textbf{97.20} & \textbf{96.19} & \textbf{98.11} \\
        Hard  & 55.1 (0.454) & 78.01 (0.183) & 82.11 & 81.72 & 78.92 \\
        \bottomrule
    \end{tabular*}
    \label{tab:hard_easy}
\end{table*}

%% file: supp_sec/experiments/easy_vs_hard_vis.tex
\begin{figure}[t]
    \centering
    \includegraphics[width=\linewidth]{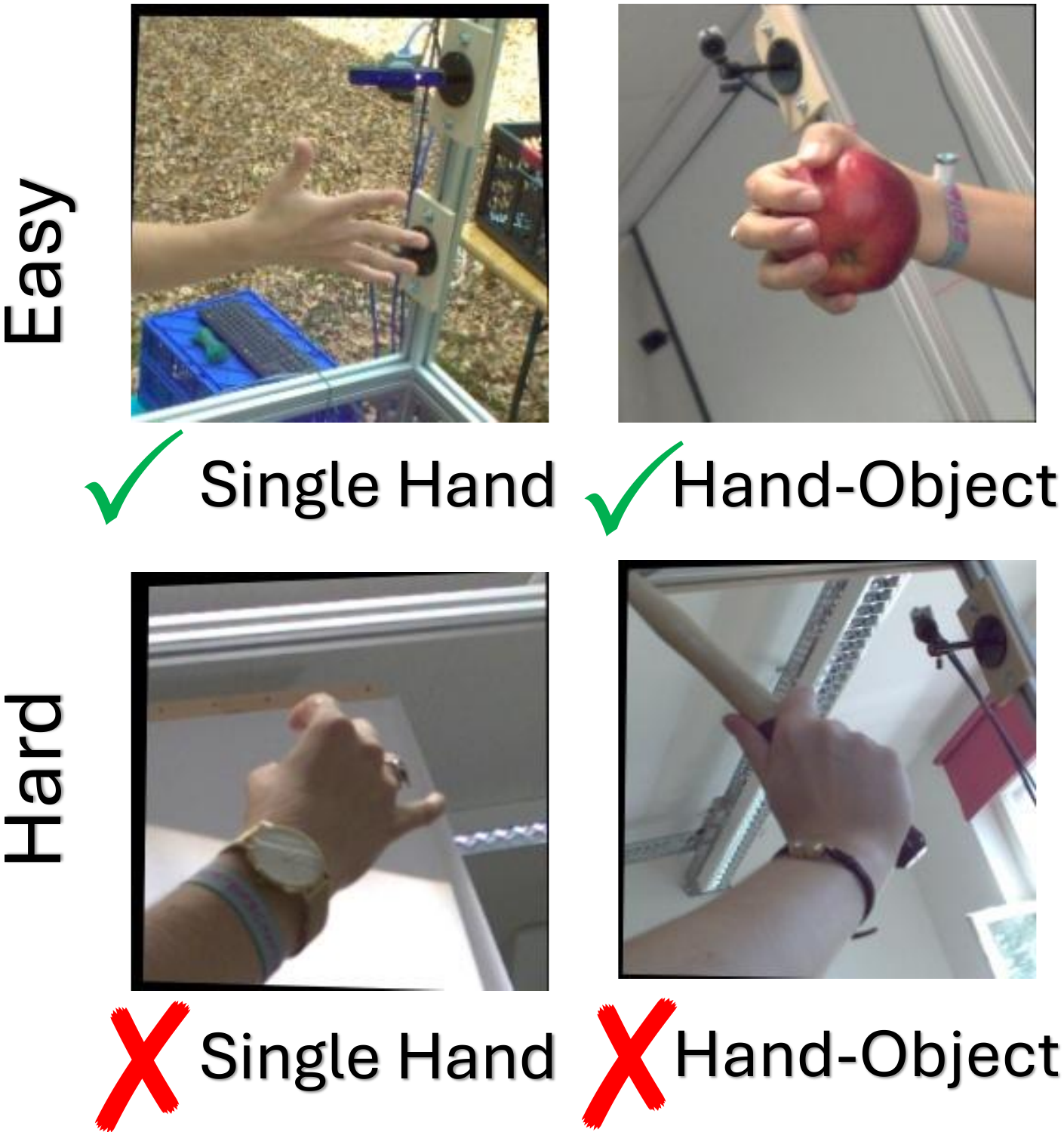}
    \caption{Easy vs Hard Examples Performance of LLaVA Mistral 7B \textit{finetuned}. Correct(\textcolor{TealBlue}{\checkmark})/wrong(\textcolor{red}{X}) QA samples for ``Which is the distance between thumb tip and index fingertip?''.}
    \label{fig:easy_vs_hard}
\end{figure}

%% file: supp_sec/experiments/human_eval.tex
\begin{table}[t!]
\centering
\caption{Human vs VLMs Accuracy on a small subset of HandVQA.}
\label{tab:human_vs_vlm}
\begin{tabular}{lc}
\toprule
\textbf{Models} & \textbf{Overall Accuracy} \\
\midrule
LLaVA Mistral 7B & 41.96\% \\
DeepSeek Janus Pro 7B & 45.82\% \\
Qwen 2.5 VL 7B & 41.97\% \\
Humans & 80.94\% \\
\bottomrule
\end{tabular}
\end{table}

%% file: supp_sec/license_details.tex
\section{License Details of Source Datasets}

\textit{HandVQA} is constructed entirely from three existing and publicly available 3D hand datasets: FreiHAND \cite{zimmermann2019freihand}, InterHand2.6M \cite{Moon_2020_ECCV_InterHand2.6M}, and FPHA \cite{garcia2018first}. Each dataset is properly cited in the main paper and used in accordance with its respective license and terms of use.

\begin{itemize}
    \item \textbf{FreiHAND} is released strictly for research purposes only. Any commercial use is explicitly prohibited, and users are required to cite the original paper if the dataset or parts of it are used.
    
    \item \textbf{InterHand2.6M} is distributed under the CC-BY-NC 4.0 license, which permits use for non-commercial research with appropriate credit to the original authors.
    
    \item \textbf{FPHA} is available for free for academic research and non-commercial use.
\end{itemize}

We have adhered to the terms and conditions of each dataset as per their official distribution policies. This ensures that all licenses are respected in full, and no proprietary or restricted-use data is included in HandVQA.

%% file: supp_sec/gesture_annotations.tex
\begin{table*}[t]
\centering
\caption{Gemini-generated descriptions for HaGRID gesture labels.}
\scriptsize
\setlength{\tabcolsep}{3pt}
\renewcommand{\arraystretch}{2}
\begin{tabular}{l p{0.4\textwidth} p{0.4\textwidth}}
\toprule
\textbf{Base Label} & \textbf{Description 1} & \textbf{Description 2} \\
\midrule
\texttt{one} &
A single index finger is extended, representing the number one. &
The hand is held up with only the index finger pointing upwards. \\

\texttt{two\_up} &
The index and middle fingers are extended upwards (peace sign), representing two. &
A ``V'' sign, often used for ``two'' or ``peace''. \\

\texttt{two\_up\_inverted} &
An inverted ``V'' sign, with the palm facing inward. &
The gesture for ``two'' or ``peace'' but inverted. \\

\texttt{three} &
Three fingers are extended to represent the number three. &
The hand gesture indicates a quantity of three. \\

\texttt{four} &
Four fingers are extended to show the number four. &
The hand gesture indicates a quantity of four. \\

\texttt{fist} &
The hand is closed tightly into a fist. &
All fingers are curled inward with the thumb wrapped around them. \\

\texttt{palm} &
The hand is open with the palm facing forward, showing five fingers. &
An open palm, often representing the number five. \\

\texttt{ok} &
The thumb and index finger form a circle to signify ``OK''. &
A hand gesture indicating that everything is alright. \\

\texttt{peace} &
The index and middle fingers form a ``V'' to symbolize peace, with the palm facing out. &
A hand gesture representing peace or victory. \\

\texttt{peace\_inverted} &
The ``V'' sign for peace is made with the palm facing inward. &
An inverted peace sign. \\

\texttt{rock} &
The index finger and little finger are extended in a ``rock on'' gesture. &
The hand forms a horns sign, often associated with rock music. \\

\texttt{hand\_heart} &
Both hands are brought together to form the shape of a heart. &
A symbol of love or affection made with the hands. \\

\texttt{like} &
The thumb is pointed up in a ``thumbs-up'' gesture of approval. &
A ``like'' or ``good job'' sign made with the thumb. \\

\texttt{dislike} &
The thumb is pointed down in a ``thumbs-down'' gesture of disapproval. &
A ``dislike'' sign made by pointing the thumb downwards. \\

\texttt{stop} &
The hand is held up with the palm facing forward to signal ``stop''. &
A universal sign to halt or cease an action. \\

\texttt{stop\_inverted} &
An inverted version of the stop gesture. &
A stop sign made with the back of the hand facing forward. \\

\texttt{point} &
The index finger is extended to point at a person or object. &
A gesture used to indicate a direction or draw attention to something. \\

\texttt{grabbing} &
The hand is held up with fingers spread and curled, as if grabbing a large object. &
A claw-like gesture used to represent grabbing. \\

\texttt{grip} &
Holding a small object securely between the thumb and index finger. &
A precision grip used to manipulate a small item. \\

\texttt{call} &
The thumb and little finger are extended in a ``call me'' gesture. &
The hand shape mimics holding a telephone receiver. \\

\texttt{timeout} &
The hands form a ``T'' shape, signaling a pause or timeout. &
A common gesture in sports to request a break. \\

\texttt{no\_gesture} &
The hand is in a neutral, resting state with no specific gesture. &
No specific gesture is being performed by the hand. \\

\texttt{holy} &
The hands are held together in a prayer-like or reverent gesture. &
A gesture symbolizing prayer or respect. \\

\texttt{little\_finger} &
Only the little finger is extended, often for a promise or pinky swear. &
The pinky finger is held up. \\

\texttt{middle\_finger} &
An offensive gesture with the middle finger extended. &
The middle finger is raised while the others are in a fist. \\

\texttt{mute} &
A gesture indicating a request for silence. &
The hand covers the mouth or fingers are held to the lips to mean ``be quiet''. \\

\texttt{take\_picture} &
The hand mimics the action of pressing a camera shutter button. &
A gesture that looks like someone is taking a photograph. \\

\texttt{three\_gun} &
A gesture resembling a gun, often made with the thumb and first two fingers. &
A playful hand gesture shaped like a firearm. \\

\texttt{thumb\_index} &
A single hand is held up with the thumb and index finger extended to form an ``L'' shape. &
A one-handed gesture shaped like the letter ``L''. \\

\texttt{thumb\_index2} &
Both hands simultaneously form an ``L'' shape with the thumb and index finger. &
A two-handed gesture where each hand makes the shape of the letter ``L''. \\

\texttt{xsign} &
The arms are crossed over the chest to form an ``X''. &
A defensive or blocking gesture made by crossing the arms. \\
\bottomrule
\end{tabular}
\label{tab:hagrid_descriptions}
\end{table*}

%% file: supp_sec/figures/qualitative_results_gesture.tex
\begin{figure*}[t!]
\centering
\begin{adjustbox}{max width=\textwidth, max height=\textheight}
  \includegraphics{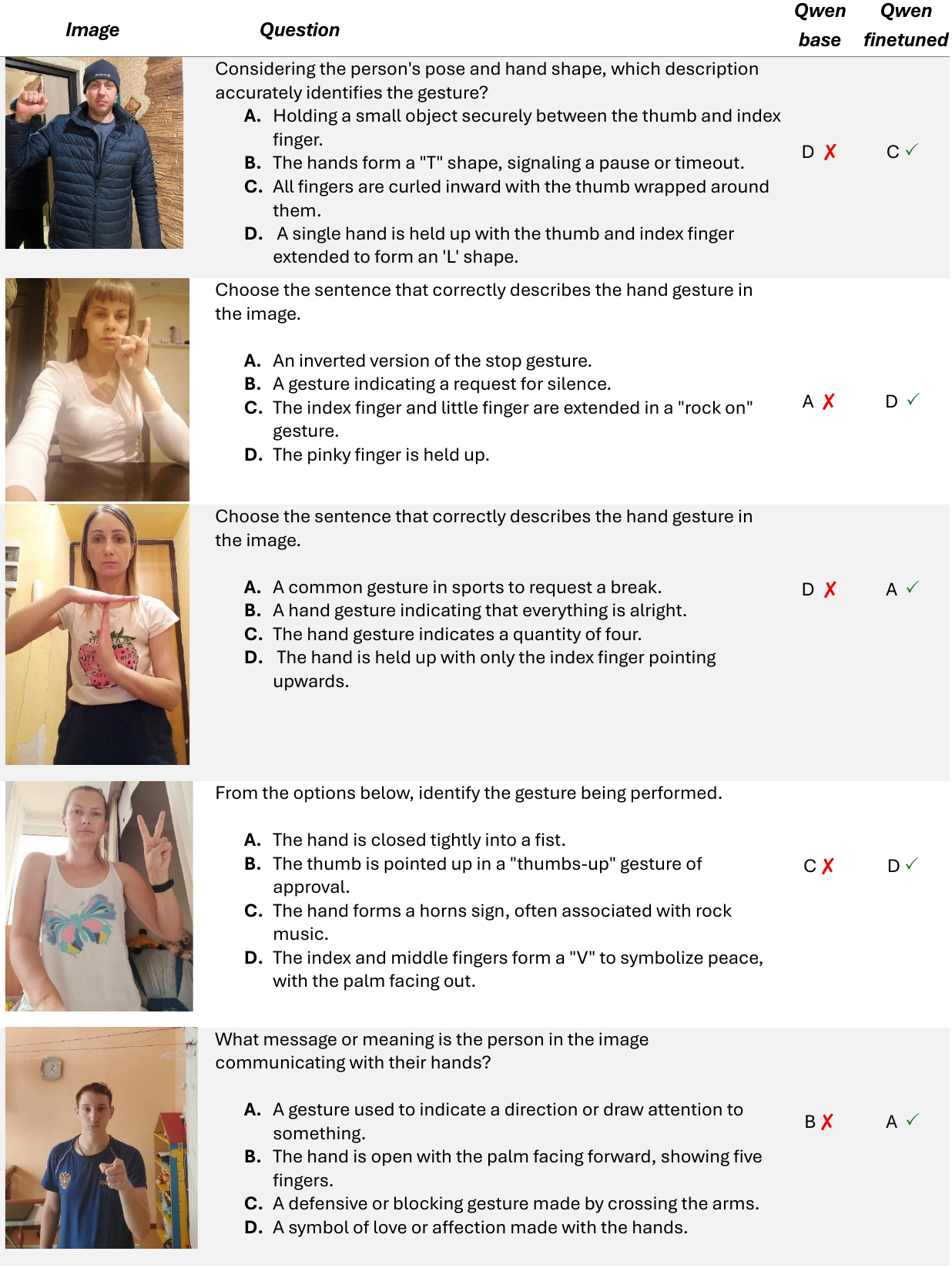}
\end{adjustbox}
\caption{ \textbf{Qualitative Results on Zero-shot Gesture Recognition on HaGRID dataset \cite{Kapitanov_2024_WACV}.}}
\label{fig:qualitative_gesture}
\end{figure*}

%% file: supp_sec/figures/qualitative_results_interaction.tex
\begin{figure*}[t!]
\centering
\begin{adjustbox}{max width=\textwidth, max height=\textheight}
  \includegraphics{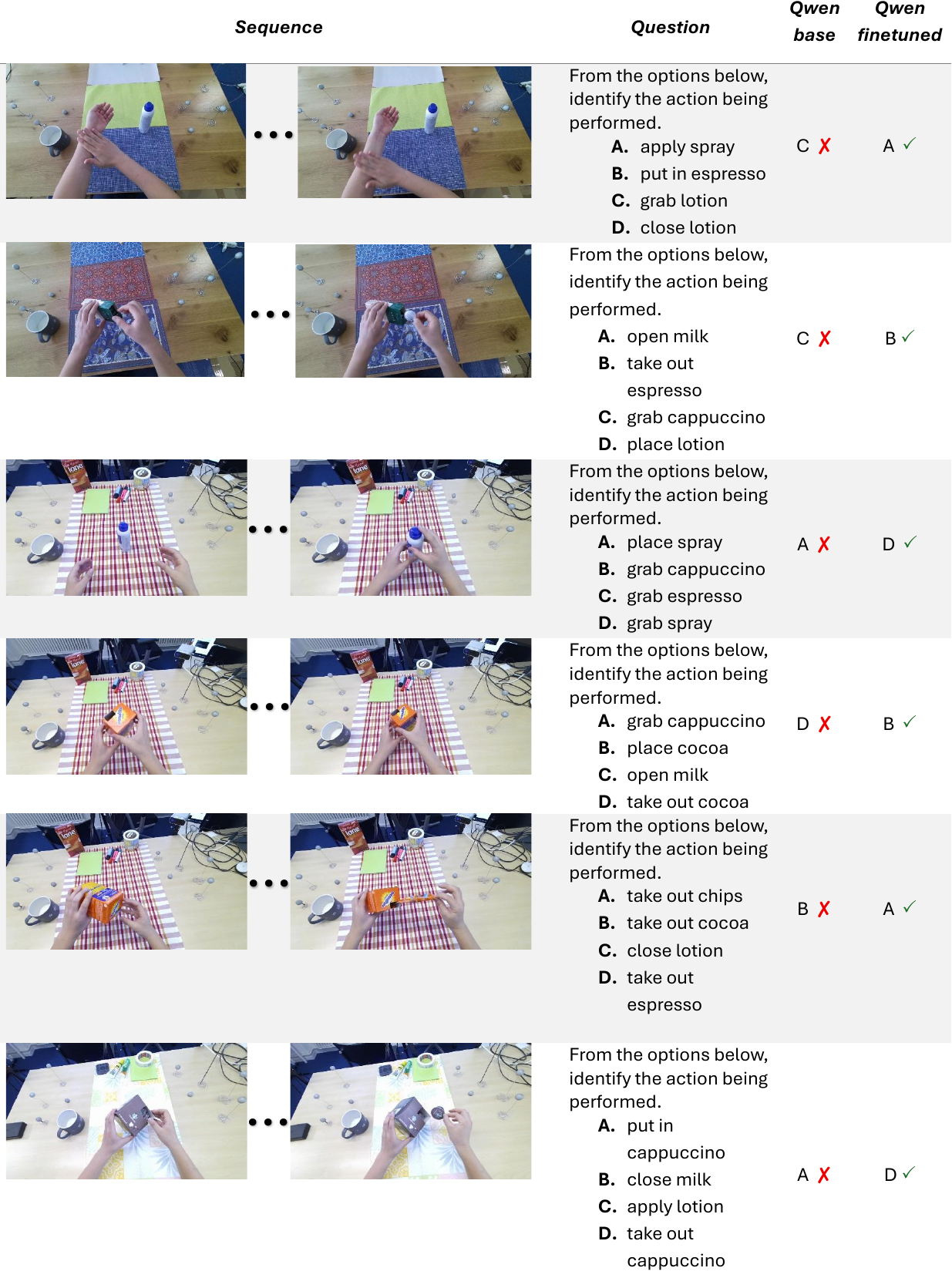}
\end{adjustbox}
\caption{ \textbf{Qualitative Results on Zero-shot Hand-Object Interaction Recognition on H2O dataset \cite{Kwon_2021_ICCV}.}}
\label{fig:qualitative_interaction}
\end{figure*}

%% file: supp_sec/figures/qualitative_results_freihand.tex
\begin{figure*}[t!]
\centering
\begin{adjustbox}{max width=\textwidth, max height=0.95\textheight}
  \includegraphics{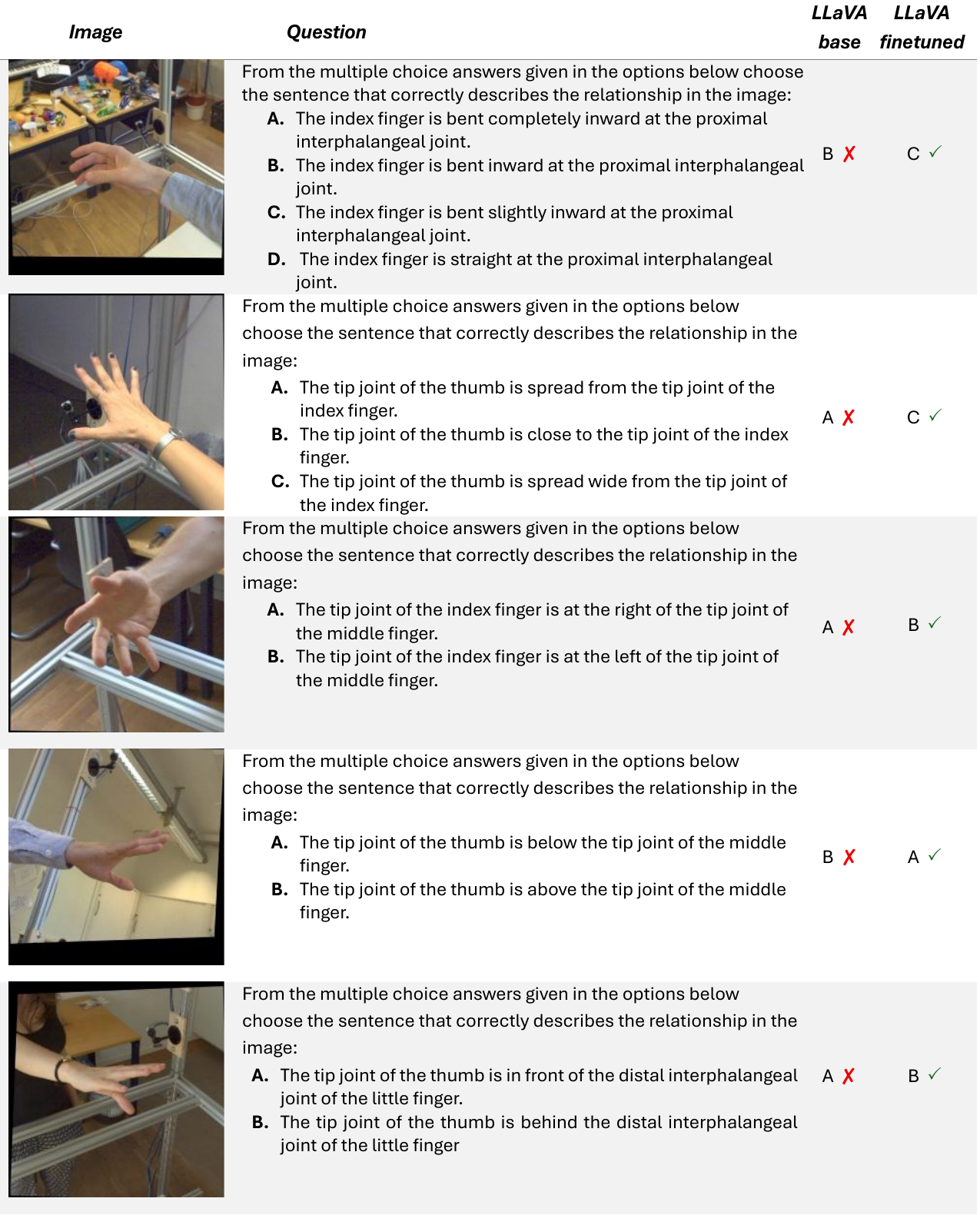}
\end{adjustbox}
\caption{ \textbf{Qualitative Comparison on FreiHAND \cite{zimmermann2019freihand}.}
Examples comparing LLaVA (base) and LLaVA fine-tuned on FreiHAND.} 
\label{fig:qualitative_freihand}
\end{figure*}

%% file: supp_sec/figures/qualitative_results_interhand.tex
\begin{figure*}[t!]
\centering
\begin{adjustbox}{max width=\textwidth, max height=0.95\textheight}
  \includegraphics{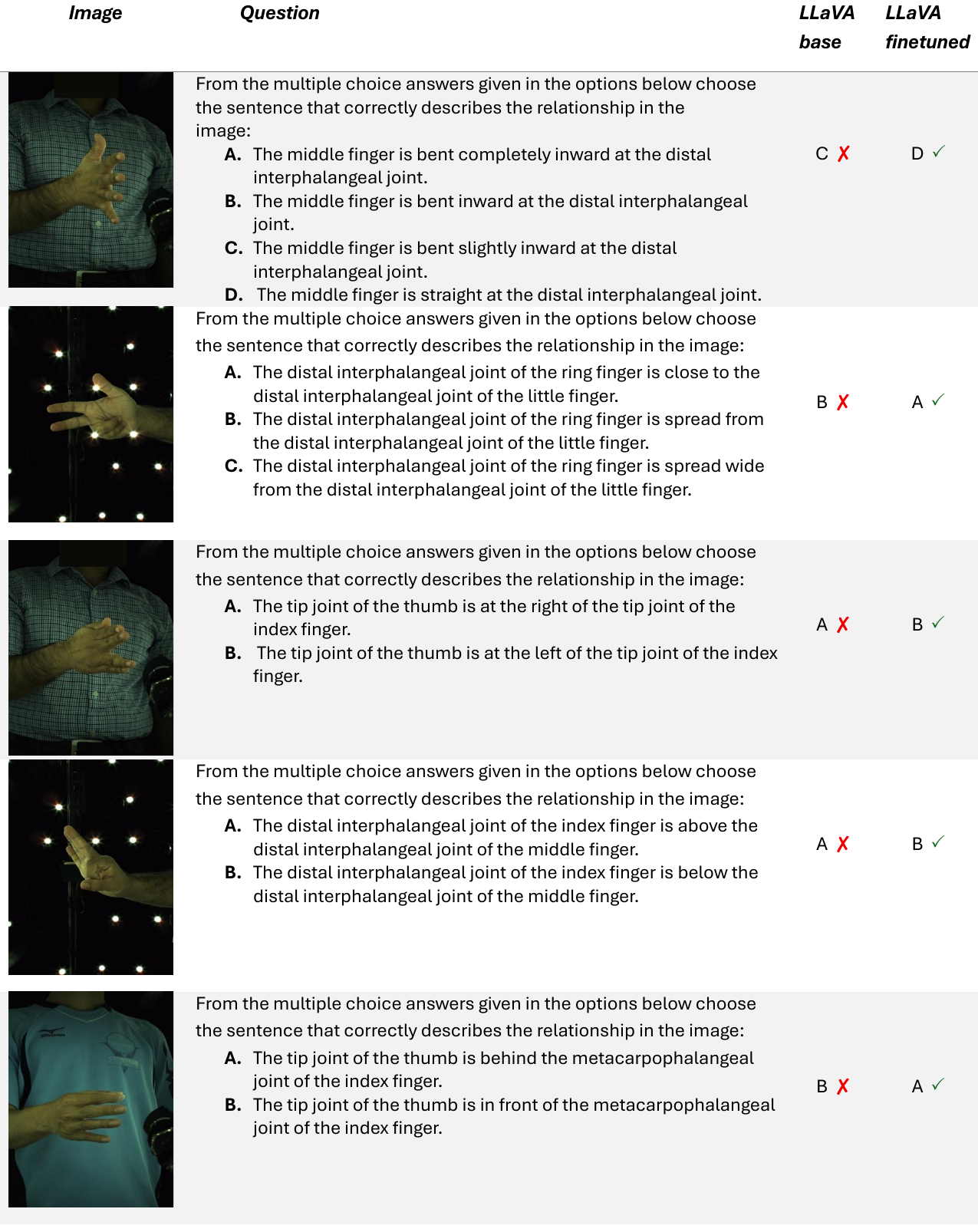}
\end{adjustbox}
\caption{\textbf{Qualitative Comparison on InterHand2.6M \cite{Moon_2020_ECCV_InterHand2.6M}.}
Examples comparing LLaVA (base) and LLaVA fine-tuned on InterHand2.6M.} 
\label{fig:qualitative_interhand}
\end{figure*}

%% file: supp_sec/figures/qualitative_results_fpha.tex
\begin{figure*}[t!]
\centering
\begin{adjustbox}{max width=\textwidth, max height=0.95\textheight}
  \includegraphics{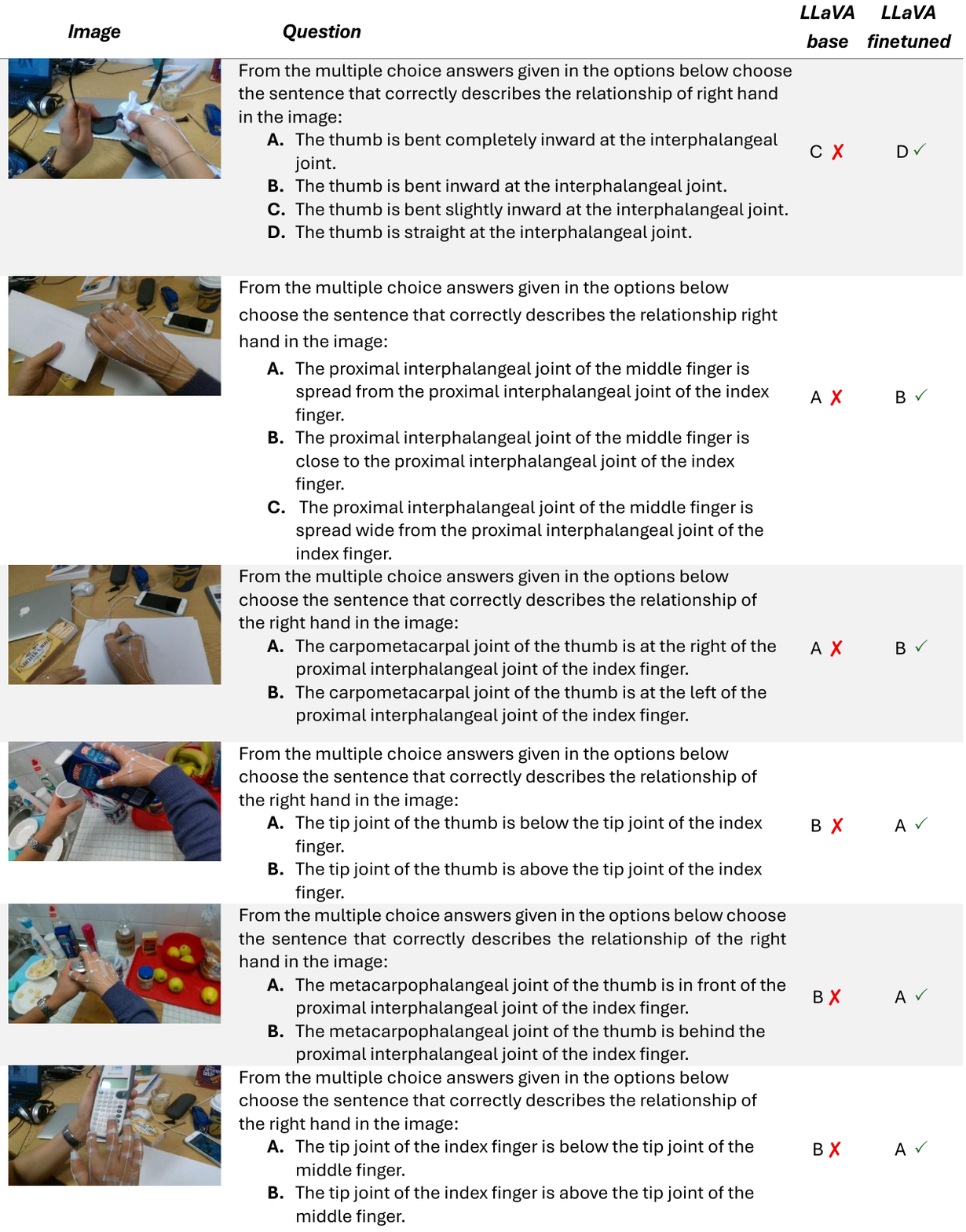}
\end{adjustbox}
\caption{\textbf{Qualitative Comparison on FPHA \cite{garcia2018first}.}
Examples comparing LLaVA (base) and LLaVA fine-tuned on FPHA.} 
\label{fig:qualitative_fpha}
\end{figure*}

%% file: supp_sec/figures/qualitative_results_ood.tex
\begin{figure*}[t!]
\centering
\begin{adjustbox}{max width=\textwidth, max height=0.95\textheight}
  \includegraphics{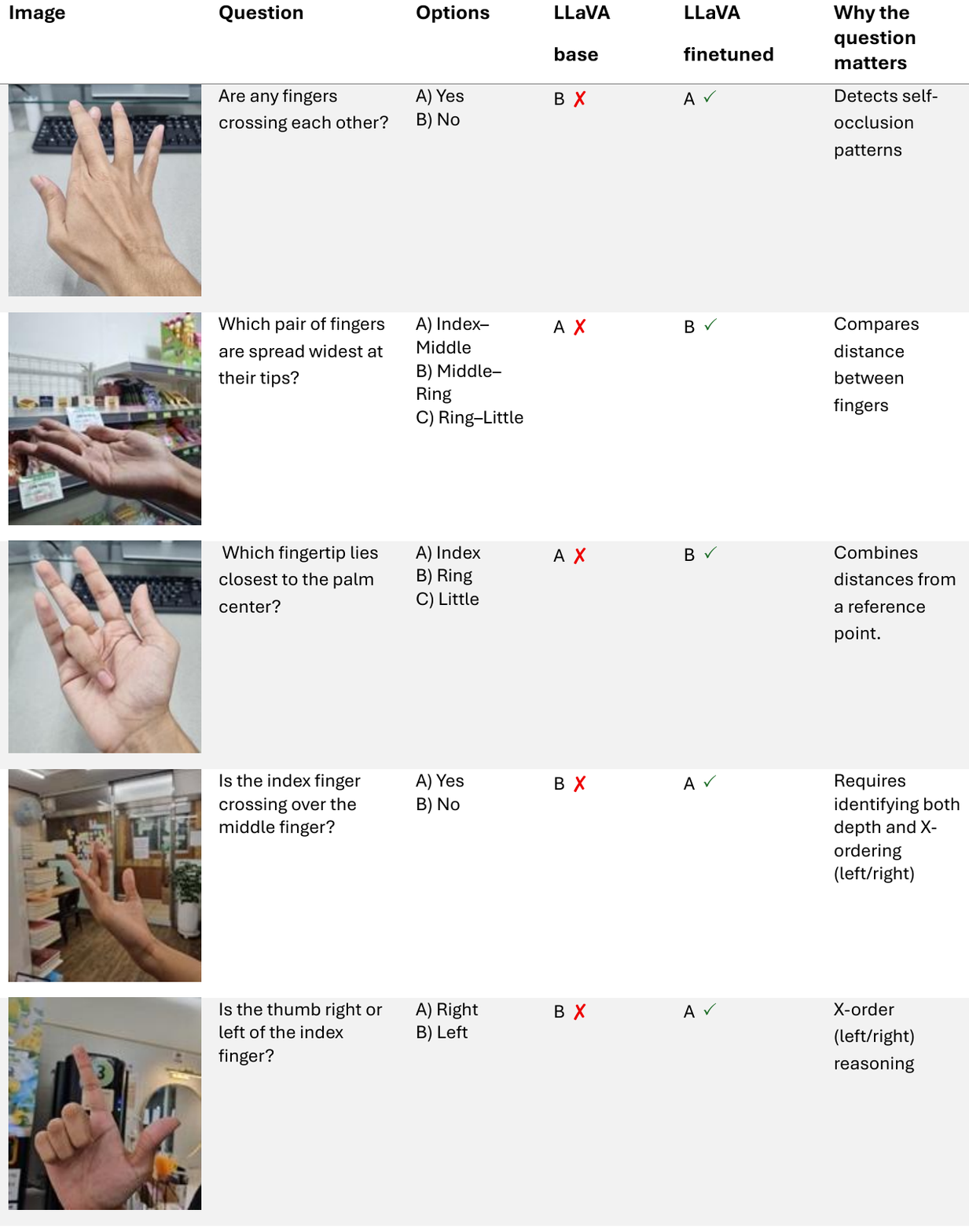}
\end{adjustbox}
\caption{ \textbf{Qualitative Results on In-the-Wild Images.}
We evaluate spatial reasoning on challenging questions using in-the-wild images. The fine-tuned LLaVA outperforms the base model on tasks involving occlusion, depth, and inter-finger relationships, demonstrating improved generalization beyond the training data.}
\label{fig:qualitative_ood}
\end{figure*}